\documentclass[sigconf, balance=false, authorversion, nonacm]{acmart}

\usepackage[utf8]{inputenc}
\usepackage{lipsum}
\usepackage[group-separator={,}]{siunitx}
\usepackage{tikz}
\usetikzlibrary{matrix, positioning, fit, plotmarks, patterns, tikzmark, patterns.meta}
\usepackage{pgfplots}
\usepgfplotslibrary{statistics,groupplots}
\usepackage{colortbl}
\usepackage{subfig}
\usepackage{makecell}
\usepackage{amsmath}
\usepackage{xspace}
\usepackage{extdash}
\usepackage{pifont}
\usepackage{graphicx}

\def\mystrut{\vphantom{hg}}
\def\titlestrut{\vphantom{hg}}

\pgfplotscreateplotcyclelist{embedding list}{
  red\\
  blue\\
  black\\
  orange\\
  violet\\
  teal\\
}

\pgfplotsset{
  compat=1.17,
  cycle list name=embedding list,
}

\title[On the Role of Pre-trained Embeddings in Binary Code
Analysis]{On the Role of Pre-trained Embeddings \\ in Binary Code
  Analysis\vspace{0.3cm}}
\author{Alwin Maier} \affiliation{ \institution{Max Planck
    Institute for \\ Solar System Research} \country{Germany} }
\author{Felix Weißberg} \affiliation{ \institution{Technische Universität Berlin}
  \country{Germany} \vspace{0.8cm} } 
\author{Konrad Rieck} \affiliation{ \institution{Technische Universität Berlin\\ \& BIFOLD}
  \country{Germany} \email{\phantom{rieck@tu-berlin.de}} }

\setcopyright{cc}
\setcctype{by-sa}

\makeatletter
\renewcommand{\@copyrightpermission}{%
  {\footnotesize  
  This is a preprint of the paper published in the proceedings of the 19th ACM Asia Conference on Computer and Communications Security (AsiaCCS). The final version is available at \href{https://doi.org/10.1145/3634737.3657029}{https://doi.org/10.1145/3634737.3657029}
  } \\}
\makeatother 

\keywords{Transfer learning, Binary code analysis}

\ccsdesc[300]{Computing methodologies~Learning paradigms}

\mathchardef\mhyphen="2D
\newcommand{\insn}{I}

\newcommand{\seq}{S}
\newcommand{\iset}{\mathcal{I}}
\newcommand{\sset}{\mathcal{S}}

\newcommand{\emb}{\Phi}
\newcommand{\embi}{\Psi}
\newcommand{\embf}{\Omega}
\newcommand{\mnem}{M}
\newcommand{\op}{O}
\newcommand{\dataset}{\mathcal{D}}

\newcommand{\palmtree}{\textsl{PalmTree}\xspace}
\newcommand{\wtov}{\textsl{Word2Vec}\xspace}
\newcommand{\itov}{\textsl{Instruction2Vec}\xspace}
\newcommand{\itovx}{\textsl{Instr.2Vec}\xspace}
\newcommand{\atov}{\textsl{Asm2Vec}\xspace}
\newcommand{\etoe}{\textsl{end\Hyphdash*to\Hyphdash*end}\xspace}
\newcommand{\rand}{\textsl{random}\xspace}
\newcommand{\ada}{\textsl{ada-002}\xspace}
\newcommand{\Gemini}{\textsl{Gemini}\xspace}
\newcommand{\SAFE}{\textsl{SAFE}\xspace}

\newcommand{\gcc}{\textsl{GCC}\xspace}
\newcommand{\clang}{\textsl{CLang}\xspace}
\newcommand{\debian}{\textsl{Debian}\xspace}

\newcommand*\circled[1]{\tikz[baseline=(char.base)]{
            \node[shape=circle,draw,inner sep=1.0pt] (char) {\small{#1}};}}

\newcommand{\performance}{\mathcal{A}}

\begin{document}

\begin{abstract}
  Deep learning has enabled remarkable progress in binary code
  analysis. In particular, pre-trained embeddings of assembly code
  have become a gold standard for solving analysis tasks, such as
  measuring code similarity or recognizing functions. These
  embeddings are capable of learning a vector representation from
  unlabeled code. In contrast to natural language processing, however,
  label information is not scarce for many tasks in binary code
  analysis. For example, labeled training data for function
  boundaries, optimization levels, and argument types can be easily
  derived from debug information provided by a compiler. Consequently,
  the main motivation of embeddings does not transfer directly to
  binary code analysis.

  In this paper, we explore the role of pre-trained embeddings from a
  critical perspective. To this end, we systematically evaluate recent
  embeddings for assembly code on five downstream tasks using a corpus
  of 1.2 million functions from the Debian distribution. We observe
  that several embeddings perform similarly when sufficient labeled
  data is available, and that differences reported in prior work are
  hardly noticeable. Surprisingly, we find that end-to-end learning
  \emph{without} pre-training performs best on average, which calls
  into question the need for specialized embeddings. By varying the
  amount of labeled data, we eventually derive guidelines for when
  embeddings offer advantages and when end-to-end learning is
  preferable for binary code analysis.
\end{abstract}

\maketitle

\section{Introduction}

Deep learning has been a driving force behind several advances in
computer security. In particular, the ability of neural networks to
distill information from highly complex data, such as assembly code,
has led to a number of learning-based methods for binary code
analysis. These methods allow, for example, to locate function
boundaries~\citep{PeiGuaWil+21, AlvSon19, ShiSonMoa15}, differentiate
optimization levels~\citep{PizIno21, CheShiLi+18}, assess code
similarity~\citep{XuLiuFenYin+17, ZuoLiYouLuo+19, MasAntPet+19,
  RedLuoZen19}, reconstruct arguments~\citep{ChuSheSaxLia+17,
  JinPeiWonLin+22, HeIvaTsaRay+18}, and detect aliases in memory
\citep{GuoMuXin19}.  While the approaches differ in the architecture
of the neural networks used, most share a key component: an
\mbox{\emph{embedding}}. This learned vector representation originates
from the area of natural language processing and provides geometric
access to the data's structure and semantics, forming a versatile
basis for solving different learning tasks.

Over recent years, several methods have emerged for crafting
embeddings tailored to binary code analysis, including
\atov~\citep{DinFunCha19}, \itov~\citep{LeeKwoChoLimBaePar19}, and
\palmtree~\citep{LiQuYin21}. Additionally, dedicated approaches such
as \Gemini~\citep{XuLiuFeng+17} and \SAFE~\citep{MasLunPet+19} have
been specifically designed for generating function embeddings used to
detect similar functions.
The underlying rationale for these embeddings lies in their
\emph{pre-training} on large collections of unlabeled code, which
allows for encoding general characteristics the data and learning a
versatile representations for various downstream tasks.
Over time, these embeddings have become a gold standard for applying
deep learning to binary code analysis~\citep{AhnAhnKooPae+22,
  BisBarLaz+22, PeiGuaBroChe+21, JinPeiWonLin+22, YuRuiQiy20,
  ZuoLiYouLuo+19}.

Although natural language processing bears
similarities with
code analysis, the availability of labeled data differs
between the two domains. For natural language text, tedious manual
labeling is often unavoidable to create examples for supervised
learning, rendering pre-trained embeddings indispensable.
In contrast, for many tasks of binary code analysis, labeled data can
be easily generated from debug information provided by a compiler. For
example, labels for function boundaries, optimization levels, and
argument types can be extracted during the compilation process and
enable constructing large-scale training sets with label information
automatically.
As a result, the necessity of pre-trained embeddings in natural
language processing does not naturally apply to tasks in binary code
analysis, where end-to-end learning is often possible.

In this paper, we investigate the role of pre-trained embeddings
for binary code from a critical perspective. For this investigation,
we construct a labeled evaluation corpus of 1.2 million functions from
the Debian distribution, totaling about 129 million x86
instructions. This corpus allows us to systematically evaluate the
capabilities and limitations of five widely used embeddings for
assembly code, namely \wtov~\cite{MikSutCheCorDea13},
\atov~\citep{DinFunCha19}, \itov~\citep{LeeKwoChoLimBaePar19}, and
\palmtree~\citep{LiQuYin21}. In particular, we evaluate the
performance of each embedding in different configurations on five
common downstream tasks of binary code analysis: compiler detection,
optimization level identification, function argument prediction,
argument type reconstruction and code similarity detection.

Our results provide a new view on pre-trained embeddings in binary
code analysis: First, we observe that the embeddings hardly differ in
performance if sufficient training data is available,
so that differences discussed in prior work are not
noticeable~\citep{LiQuYin21}. Even a random instruction embedding
provides a reasonable performance in our
experiments. Second, we surprisingly find that end-to-end learning
\emph{without} a pre-trained embedding yields the best performance on
average. Contrary to our intuition, pre-training does not generally
unlock additional information, and it provides no advantage in binary
code analysis when sufficient labels are available.

Our work should not be interpreted as a general criticism of
pre-trained embeddings: By reducing the amount of labeled
data, we can also demonstrate the utility of this technique
in our evaluation. In particular, PalmTree~\citep{LiQuYin21} provides
the best overall performance when labeled training data becomes
scarce. We can derive guidelines to help practitioners decide
whether or not to use pre-training.
Consequently, our work adds a new facet to research on deep learning
for binary code analysis, indicating that benefits from other domains
do not necessarily carry over and need be critically reflected.
In summary, we make the following contributions:
\vspace{0.2cm}
\begin{enumerate}\setlength{\itemsep}{3pt}
  
\item \emph{Critical evaluation of pre-trained embeddings.} We present
  a systematic evaluation of embeddings for binary code analysis with
  varying training data and embedding dimensions while also
  considering computational expenses.
  
\item \emph{Large-scale evaluation corpus.} We provide researchers
  with an open corpus of 1.2~million labeled functions and 129~million
  x86~instructions from the Debian distribution for five
  downstream tasks (\url{https://github.com/a0x77n/orbit-dataset}).
  
\item \emph{Recommendations for binary code analysis.} We derive
  guidelines to study the performance of pre-trained embeddings on
  assembly code and decide when to rely on conventional end-to-end
  learning instead.

\end{enumerate}

\paragraph{Roadmap.} We briefly review the background of pre-training
and embeddings for assembly code in
Section~\ref{sec:pre-training}. Our benchmark corpus of labeled code
for five downstream tasks is then introduced in
Section~\ref{sec:benchmark} and the corresponding experiments in
Section~\ref{sec:experiments}. We discuss our findings on pre-trained
embeddings and derive recommendations in
Section~\ref{sec:results}. Finally, Section~\ref{sec:conclusion}
concludes the paper.

\section{A Primer on Pre-Training}
\label{sec:pre-training}

Let us start by introducing some background on the concept of
pre-training and embeddings for assembly code, before critically
reflecting on their role in binary code analysis.

\subsection{Training with a Headstart}
\label{sec:pre-training-vs}

When humans learn a new task, such as a playing a music instrument,
they usually do not start from scratch but are able to built upon
prior knowledge. For example, someone who played the violin is
probably faster in learning to play the cello than someone without
prior experience. This insight into the human learning process has
also been employed in the machine-learning domain and is commonly
referred to as transfer learning.

To be more specific, a learning model trained for a certain task can
help create a second model for a different task.
The first task is commonly referred to as \emph{pre-training task} and
the second as \emph{downstream task}.  The rationale underlying this
transfer is that the second model may perform better when building on
the knowledge of the first one.  Since this transfer learning
typically revolves around improving the performance of downstream
task, we refer to the pre-training as task-agnostic (with respect to
the downstream task), while the downstream training is task-specific.

\begin{figure}[htbp]
  \centering \subfloat[Pre-training and downstream task]
  {\includegraphics[width=0.98\columnwidth]{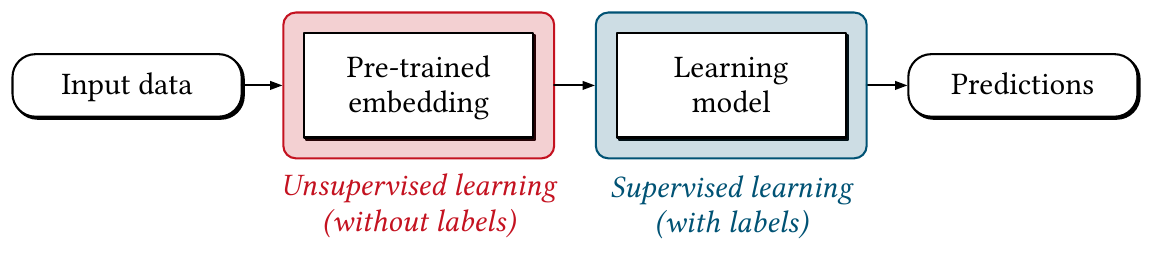}}

  \subfloat[End-to-end learning]%
  {\includegraphics[width=0.98\columnwidth]{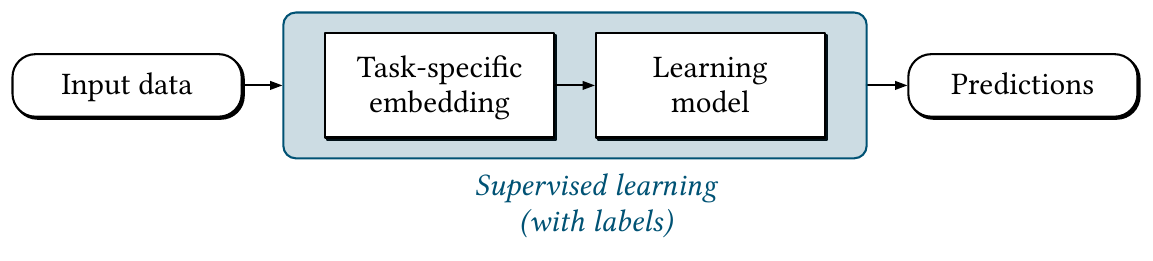}}

  \caption{Schematic comparison of pre-training task with downstream
    task and conventional end-to-end learning.\vspace{0.2cm}}
  \label{fig:overview}
\end{figure}

Pre-training is considered especially beneficial in cases where high
quality labelled data is expensive to create but unlabelled data is
widely available. In natural language processing and computer vision,
for example, vast amounts of unlabeled data are available on the
Internet and can be employed for pre-training.
Since the pre-training task and the downstream task do not need to
share the same training objective, it becomes possible to use
unsupervised learning for pre-training on a large unlabeled dataset
and perform supervised learning on a smaller labeled one.
A well-known example of this strategy is the unsupervised learning of
input representations. In this case, the pre-training task learns a
vector representation of the data, denoted as \emph{embedding}, which
serves as input for the subsequent downstream tasks, as shown in
Figure~\ref{fig:overview}(a).

This type of pre-training has been shown to be effective for several
learning tasks in several domains. For example, in computer vision,
pre-training has been employed for object
recognition~\cite{DonJiaVin14} and semantic
segmentation~\cite{LinCheCoh17}. Similarly, in natural language
processing, embeddings have been successfully applied for question
answering~\cite{DevChaLeeTou19}, machine
translation~\cite{EduOttAul18}, and text
summarization~\cite{AghGupShr21}.
Naturally, these advances have spawned a research in security aimed at
improving binary code analysis. The rationale of this work has been to
create \emph{embeddings} for assembly code that provide a versatile
representation, simplifying downstream tasks, such as function
recognition and type inference.

Pre-training contrasts with so-called \emph{\etoe learning}. In this
setup, the representation of the data is learned along with the task
at hand, as shown in Figure~\ref{fig:overview}(b). The previously
separate tasks now focus on the same objective and work with the same
labeled data. That is, the learned representation becomes
task-specific and is only suitable for the particular downstream
task. Unlabeled data cannot be used in this setting. Consequently,
\etoe learning requires access to a sufficient amount of labeled data.

\subsection{Instruction Embeddings}
\label{sec:instr-embedd}

We proceed to investigate the different embeddings that have been
proposed for binary code analysis. To this end, we first give a formal
definition and then provide a recap of recent approaches for embedding
assembly code.

Simply put, embeddings are functions mapping data of the input domain
to a vector representation. This vector can then be used for
subsequent machine-learning tasks. The main reason behind using such a
representation is, that it does not simply encode a word as a
low-dimensional vector but is also able to express relationships among
different words. For example, synonyms may be mapped to vectors within
close proximity, while opposites are a mapped far apart from each
other. Since these are appealing characteristics for machine learning,
embeddings are widely used for natural language tasks
\cite{DevChaLeeTou19, EduOttAul18, AghGupShr21} and recently for
binary code analysis, where instructions take up the role of words.

\begin{figure}[h]
  \includegraphics[trim={4.7cm 21.35cm 8cm 4.5cm},clip,width=0.95\columnwidth]{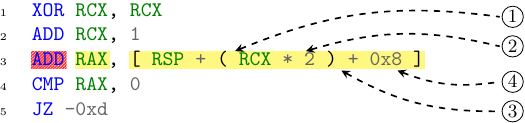}
  \vspace{-0.25cm}
  \caption{\vspace{-0.3cm}Example of x86-64 assembly instructions.\vspace{0.1cm}}
  \label{fig:asm}
\end{figure}

Before introducing these embeddings, we first need to formally define
what an instruction is. In particular, we refer to an instruction as
the disassembled representation of a machine code operation for a
given architecture. Generally, an instruction consists of an operation
type denoted as \emph{mnemonic} ($\mnem$) and a group of
\emph{operands} \mbox{($\op_1$, \dots, $\op_n$)}. Typically, the
operand of an instruction can have one of the following types:
\begin{enumerate}
  \setlength{\itemsep}{2pt} 
\item \emph{Register.} The operand is a processor \emph{register},
  where each register has a unique identifier.

\item \emph{Immediate value.} The operand is an address value or a
  numerical constant.
  
\item \emph{Memory access.} The operand is a memory access expression
  which usually consists of several components.

\end{enumerate}
While the specific structure of the operands and the instruction
depends on the architecture and the syntax flavor, the general
structure remains the same. To give an example for a specific case,
Figure~\ref{fig:asm} shows a sequence of five instructions for the
x86-64 architecture in Intel syntax. For the instruction in line
three, the mnemonic is highlighted in \tikzmark{red}red and the
operands in \tikzmark{yellow}yellow. The first operand is a register
operand, while the second one is a memory access expression. In this
specific case, \circled{1}, \circled{2}, \circled{3} and \circled{4}
point to the components of the memory access: the \emph{base
  register}, \emph{index register}, \emph{scale} and
\emph{displacement value}, respectively. Although these instructions
actually describe program semantics, it is easy to see that they can
be interpreted as words (token) or phrases, similar to natural
language text.

Embeddings for binary code differ in how they characterize the content
of the instructions. While most embeddings build on tokenization
schemes for operations and operands that are architecture independent,
some approaches utilize specific knowledge about instruction semantics
and execution behavior. The embeddings considered in our work fall
into the first category, with the exception of \itov, which
incorporates knowledge about the structure of x86-specific memory
access expressions.

To describe common instruction embeddings jointly, we phrase the
previous description more formally. To this end, let $\iset$ be the
set of all
instructions. The embedding method can then intuitively be described
as a function $\embi$ mapping an instruction $\insn \in \iset$ to a
real valued vector of dimension $d$:
\[
  \embi: \iset \longrightarrow \mathbb{R}^d
\]
Although sufficient for most cases, this definition is not able to
capture methods which create context-dependent embeddings, such as
\palmtree. That is, for a sequence of instructions
$\seq = (\insn_1, \dots, \insn_m)$, the embedding for one instruction
also depends on other instructions within the same sequence. To also
capture such embedding methods we extend our definition to sequences
of instructions.  Let $\sset$ be the set of all instruction sequences
$\sset = \bigcup_i \iset^i$.  This allows for the following embedding
function definition
\[
  \emb: \sset \longrightarrow \bigcup_i \mathbb{R}^{d \times i}
\]
where a whole instruction sequence is mapped to a matrix in which each
column represents the embedding vector of the respective instruction.
With this definition it is possible to map the same instruction to
different vector representations dependent on the surrounding
instructions.

The remainder of this section introduces the six instruction
embeddings used in our evaluation.

\paragraph{(a) \wtov} This embedding \cite{MasLunPet+19} is a variant
of the classic embedding introduced by \citet{MikSutCheCorDea13},
which has been originally designed for natural language text.  To
learn a vector representation of assembly instructions, one can
directly transfer this idea to binary code by considering instructions
sequences as sentences and instructions themselves as words.
The resulting embedding is based on the word embedding function $\phi$
of the original model and yields the function
\begin{equation*}
  \embi(\insn) = 
  \begin{cases}
    \phi(\insn) \text{, if $\insn$ is in the vocabulary} \\
    0^d \text{ otherwise.}
  \end{cases}
\end{equation*}
That means out-of-vocabulary (OOV) tokens are mapped to the
zero~vector.  For the considered \wtov based embeddings approach, full
instructions are considered as a single token and no further
tokenization is conducted. However, immediate values greater than
\num{5000} are normalized, i.e. replaced by the word \texttt{IMM}.
In line with \citet{MasLunPet+19}, our implementation uses the
skip-gram algorithm~\cite{MikCheCorDea13} with negative sampling
\cite{MikSutCheCorDea13}. A similar instruction embedding based on a
Word2Vec model has also been used by \citet{ChuSheSaxLia17} for
learning function type signatures.

\paragraph{(b) \itov}

The \itov embedding \cite{LeeCho+17} utilizes a classic model as
well. As opposed to \wtov, it is trained over individual instruction
components gathered from instruction sequences. Based on the
component-based embedding $\phi$ that maps a component to a $m$
dimensional space, an embedding for instructions is assembled. The
embedding vector of an instruction $\insn$ can be divided into 9~slots
of size $m$. Slot~1 contains the embedding vector of the
operation~($\mnem$), slot~2 up to slot~5 contains the embedding of the
first operand~($\op_1$), and slot~6 to slot~9 contains the embedding
of the second operand~($\op_2$). If an operand is absent, the
respective slots are set to $0^m$.
The layout of the slots of the operands depends on the type of the
respective operand:
\begin{enumerate}
\item $\op_i$ is a register $x$: $\textrm{slot}_{1 + 4*i + 1} = \phi(x)$
\item $\op_i$ is a immediate $x$: $\textrm{slot}_{1 + 4*i + 2} = (0, \dots 0, x)$
\item $\op_i$ is a memory access $(a, b, c, d)$:\\
  $\textrm{slot}_{1 + 4*i + 1} = \phi(a)$, where $a$ is the base register,\\
  $\textrm{slot}_{1 + 4*i + 2} = (0, \dots, 0, d)$, where $d$ is the displacement value,\\
  $\textrm{slot}_{1 + 4*i + 3} = \phi(b)$, where $b$ is the index register,\\
  $\textrm{slot}_{1 + 4*i + 4} = \phi(c)$, where $c$ is the scale
\end{enumerate}

The final embedding vector of $\insn$ is obtained by concatenation of
the nine slots:
$\embi(\insn) = \mathrm{slot}_1 \mathbin\Vert \dots \mathbin\Vert
\mathrm{slot}_9$. Hence, the \itov embedding has the dimension
constraint $d = 9m$. As can already be seen by the way in which the
embeddings are created, \itov uses a finer granularity compared to the
\wtov approach. That is, the instruction is broken down into mnemonics
and operands, including registers, immediates, and memory access
elements:
For example, the third instruction of Figure~\ref{fig:asm} would be
tokenized into \texttt{ADD RAX RSP RCX 2 0x8}.

\paragraph{(c) Asm2Vec} \citet{DinFunCha19} introduces Asm2Vec
\cite{DinFunCha19}, an embedding technique for binary functions based
on the document embedding method PV-DM~\cite{LeMik14}. Similar to the
\itov approach, Asm2Vec utilizes an embedding function $\phi$, mapping
instruction components to an $m$-dimensional vector. To enhance the
embedding, the instructions's syntactic structure is considered:
\[
  \embi(\insn) = \phi(\mnem) \mathbin\Vert \frac{1}{N} \sum_{n=1}^N
  \phi(\op_n)
\]
Each instruction embedding is defined as the concatenation of the
mnemonic and the averaged embeddings of the operands. The tokenization
in \atov is identical to \itov and the embedding dimension $d$
is constrained to even numbers: $d = 2m$. Asm2Vec provides embeddings
for functions and intermediate embeddings for instructions
We investigate both types in our evaluation.

\paragraph{(d) \palmtree} This embedding \cite{LiQuYin21} is the most
recent assembly language model build on top of BERT
\cite{DevChaLeeTou19}. Since BERT is based on the transformer
architecture \cite{ShaParUsz+17} its models are capable of producing
different embeddings for the same word dependent on the word's
context. \palmtree can do the same for instructions. Hence, different
from previous approaches, this embedding is context-dependent. Also
different from the previous approaches, \palmtree interprets
instructions as sentences and instruction tokens as words. Compared to
the other approaches, \palmtree uses an even more fine grained
tokenization which considers every syntactic element of the
instruction. For example, the third instruction in
Figure~\ref{fig:asm} is tokenized into \texttt{ADD RAX , [ RSP + ( RCX
  * 2 ) + 0x8 ]}.

To train the language model, two tasks from the BERT approach are
reused, namely, masked language modeling and an adapted version of
next sentence prediction. Also a third task is taken into account
which builds on the clearly documented semantics of instructions
(i. e. source and destination are known) and aims at predicting
def-use relationships between two instructions. The \palmtree
embedding function $\emb_{\text{PalmTree}}$ maps a sequence of
instructions to a real valued matrix by performing a mean pooling of
the hidden states of the second last layer of the transformer
encoder. This gives the instruction-sequence embedding function
\[
  \emb(\seq) = \emb_{\text{PalmTree}}(\seq)
\]
and a context-dependent embedding representations for the $i$-th
instruction of this sequence by $\emb_{\text{PalmTree}}(\seq)_i$.

\paragraph{(e) Random} As a baseline, we also consider a random
embedding. This embedding is conceptually trivial and just uses a
random projection to map instructions to real-valued vectors. It can
be expressed as $\embi(\insn) = r_\insn$ with $r_\insn$ being
uniformly sampled from $\mathbb{R}^d$ for each instruction. Note that
this embedding can be considered a lower bound for the utility of
embeddings, as it randomly represents instructions in a vector space
without any information inferred from real-word binary code.

\paragraph{(f) End-to-end} The conceptual antagonist to these
pre-trained embeddings is \etoe learning. In this case, a so-called
embedding layer is added to the neural network used as learning
model. This layer executes the mapping $\embi$ from tokens to vectors
before the data is forwarded to subsequent layers in the
model. Technically, $\embi$ is defined by an embedding matrix which
can be thought of as a lookup table. Different from pre-trained
embeddings, the entries of the embedding matrix are trainable
parameters of the neural network, realizing a task-specific
representation.

For our implementation of an embedding layer in end-to-end learning,
we process each function using \atov's tokenization scheme to split
instructions into tokens first. Based on these tokens we adapt a
vocabulary with \num{2048} features consisting of the \num{2046} most
frequent tokens and two special tokens, one used for masking and the
other represents the out-of-vocabulary (OOV) token. The OOV token
combines the less frequent tokens. Subsequently all tokens of a
function are encoded using a unique number for each vocabulary token,
resulting in a sequence of numbers which are streamed into the
embedding layer. The embedding layer turns each number into a
$d$-dimensional embedding vector.

\subsection{Function embeddings}
\label{sec:function-embeddings}

In contrast to embeddings at the instruction level, function
embeddings aim to generate a comprehensive vector representation of an
entire binary function. Before we present different embeddings for
binary functions, we formally describe our notion of a binary function
and define the embedding function.

A binary function is represented by its control-flow graph and
comprising basic blocks, i.e., continuous sequences of disassembled
instructions interconnected by directed edges that signify the flow of
control. Formally, let
$C = (\mathcal{V}, \mathcal{N}: \mathcal{V} \to \{x: x \subset
\mathcal{V}\})$ be a control-flow graph with a set of basic blocks
$\mathcal{V}$, and a function $\mathcal{N}$ that maps each basic block
to its successors.
We can then define an embedding function using the set of all control
flow graphs $\mathcal{C}$
as 
\[
  \embf: \mathcal{C} \longrightarrow \mathbb{R}^d
\]
that maps an arbitrary control-flow graph $C$ of a binary function to
a $d$-dimensional vector. This embedding function outputs a singular
vector representation, effectively condensing the intricate structure
of a binary function into a point within a continuous vector space,
enabling streamlined comparison and analysis of functions in binary
code.

In the following, we present four function embedding methods, two of
which leverage instruction embeddings, resulting in a total of 15
different embeddings for our evaluation.

\paragraph{(a) Asm2Vec} As highlighted in the preceding section, \atov
constitutes an embedding technique for binary functions, drawing
inspiration from the PV-DM model rooted in Word2Vec principles. In
PV-DM, the model learns to predict a word based on both context words
and a paragraph vector. Throughout the training of \atov, the vector
representations of instruction components and entire functions undergo
multiple updates. Following training, the embedding of an unknown
function $C$ is derived by applying the model through gradient
descent, allowing for the inference of vector representations
$\embf(C)$.

\paragraph{(b) Gemini} \Gemini, introduced by \citet{XuLiuFenYin+17},
employs a neural network to detect similar binary functions, building
upon the \textsl{Structure2Vec} algorithm \cite{DaiDaiSong16}. In
\Gemini, basic block-level features are iteratively aggregated into
function-level features based on the binary function's control
flow. While the original \Gemini implementation relies on manually
selected features \cite[see][]{FengZhou+16, XuLiuFeng+17} for each
basic block, we further incorporate pre-trained instruction embeddings
as well as the \etoe and \rand embedding types to generate diverse
initial representations for basic blocks.

Given the instruction sequence of a basic block
$v = (\insn_1, \dots, \insn_m) \in \mathcal{V}$ and the
instruction-sequence embedding function $\emb$, the initial
representation is computed as
\[
  x_v^\emb = \frac{1}{m} \sum_{i=1}^m \emb(v)_{\ast,i},
\]
where $\emb(v)_{\ast,i}$ is the $i$-th column and the embedding vector
of $\insn_i$. In total, we compute six different vector
representations $\embf(C)$ for each binary function
$C = (\mathcal{V}, \mathcal{N})$ using the same iterative approach of
\textsl{Structure2Vec} architecture.

\paragraph{(c) SAFE} Another method for detecting similar functions is
\SAFE \cite{MasLunPet+19}, which relies on a self-attentive neural
network. Unlike \atov and \Gemini, \SAFE operates without the need for
a control-flow graph. Instead, it treats functions as flat linear
sequences of instructions, employing a recurrent neural network. The
instruction sequence undergoes initial embedding using the \wtov
technique before being fed into the neural network. Additionally, we
incorporate the previously introduced instruction embeddings. In
total, we generate five distinct embedding types, aligning with the
network architecture proposed by \citeauthor{MasLunPet+19}. For a
given function $C = (\mathcal{V}, \mathcal{N})$ and the
instruction-sequence embedding function $\emb$, the embedding vector
$\embf(C)$ by running the embedded instruction sequence through the
neural network.

\paragraph{(d) Ada-002} Finally, we consider a large language model
for embedding. OpenAI's \emph{text-embedding-ada-002} represents their
latest model, succeeding task-specific embeddings and offering a
unified
representation\footnote{\url{https://openai.com/blog/new-and-improved-embedding-model}}. Designed
for versatile applications, including text similarity, it features a
maximum context length of \num{8192} tokens and an embedding dimension
of \num{1536}. While not exclusively tailored to binary code, it draws
strength from extensive training on a diverse corpus, making it a
state-of-the-art, all-purpose embedding.

\section{An Embedding Benchmark}
\label{sec:benchmark}

The cornerstone of our evaluation is a vast open corpus of labeled
assembly code designed for various downstream tasks. Prior to delving
into the assessment of the embeddings under consideration, we provide
an introduction to these tasks (Section~\ref{sec:downstr-tasks}) and
outline our pipeline for generating large-scale datasets automatically
to train embeddings and learn models
(Section~\ref{sec:dataset-generation}).

\subsection{Downstream Tasks}
\label{sec:downstr-tasks}

\begin{table*}
  \centering
  \caption{Number of classes and model size for each downstream
    task. The right column shows the number of additional weights for
    the embedding in the end-to-end learning setup.}
  \label{tab:tasks}
  \begin{tabular}{llrrrr}
    \toprule
    Task & Downstream application & \#\,Classes & Embedding~dim. & \#\,Total weights & \#\,Embedding~weights \\
    \midrule
    T1 & Compiler &  2 & 128 & \num{656898} (\qty{100}{\percent}) & \numproduct{262144} (\qty{67}{\percent}) \\ 
    T2 & Optimization option & 4 & 128 & \num{657412} (\qty{100}{\percent}) & \num{262144} (\qty{40}{\percent}) \\ 
    T3 & Number of parameters & 10 & 128 & \num{1350666} (\qty{100}{\percent}) & \num{262144} (\qty{19}{\percent}) \\ 
    T4 & Parameter type & 7 & 128 & \num{1349895} (\qty{100}{\percent}) & \num{262144} (\qty{24}{\percent}) \\ 
    T5 & Function similarity & 2 & 64 & -- & -- \\
    \bottomrule
\end{tabular}

\end{table*}

For our evaluation, we select five common downstream tasks by which we
measure the performance of the different embeddings.  Each task
addresses a different challenge in binary
analysis~\citep{PizIno21,CheShiLi+18,ChuSheSaxLia17,DinFunCha19,XuLiuFeng+17,MasLunPet+19}.
Note that we are not proposing new solutions for these tasks but
rather recreate existing experiments to investigate the role of
pre-trained embeddings. 
Table~\ref{tab:tasks} offers a summary of the five downstream tasks,
presenting the number of trainable parameters for the learning model
and the parameters exclusive to the embedding layer in our end-to-end
approach, along with their corresponding dimensions. Notably, Task~T5,
focused on detecting similar code, does not require an additional
learning model as it relies solely on the embedding vectors. It's
worth highlighting the substantial variation in the number of model
parameters across tasks, providing an opportunity for experimentation
with learning tasks of varying complexities in the domain of binary
code analysis.

\paragraph{Compiler and optimization options} The first two tasks deal
with the detection of compilers (Task~T1) and the identification of
optimization options (Task~T2).
Both tasks have been previously investigated by \citet{PizIno21} and
\citet{CheShiLi+18}.  Our implementation is based on the network
introduced by \citeauthor{PizIno21}.  They propose a shallow network
as learning model that consists of a single LSTM layer with a terminal
dense output layer.  The LSTM layer has an output dimension of
\num{256} and performs the actual learning. 
To handle the different classes of each task, we use
\num{2} output nodes for the compiler detection (\gcc and \clang) and
\num{4} nodes for the identification of optimization options
(\verb|O0|, \verb|O1|, \verb|O2|, \verb|O3|).
Different from the approach by \citet{PizIno21} we use disassembled
instructions as input as opposed to the raw bytes of machine code
instructions.

\paragraph{Function type signatures} The other two downstream tasks
deal with the recovery of function type signatures.  We choose the
implementation published by \citet{ChuSheSaxLia17} and reuse their
network architecture.  According to their method, we create two tasks
for this problem.  The first task predicts the number of function
parameters (Task~T3) and the second task aims at determining the data
type of the first parameter (Task~T4), such as \texttt{int} or
\texttt{char *}.

The learning model consists of three sequential GRU~layers configured
with dropouts to avoid overfitting. The final layer has \num{10} nodes
for the prediction of the number of arguments and \num{7} nodes for
the prediction of the argument types \citep[see][]{ChuSheSaxLia17}.

\paragraph{Function Similarity} In addition to assessing the efficacy
of function embeddings, we explore two distinct approaches for
learning embeddings for function similarity detection. The first
approach is based on \Gemini \cite{XuLiuFenYin+17}, which leverages a
graph neural network. Another method for identifying similar binary
functions is \SAFE~\cite{MasLunPet+19}, which is founded on a
self-attentive neural network architecture. We re-implement the
original \Gemini network and adjust parameters for \SAFE to align with
the complexity of \Gemini.

While this task doesn't inherently necessitate a learning model, as
embedding vectors can be directly compared using cosine similarity, we
train different embedding models using \Gemini and \SAFE, integrating
pre-trained instruction embeddings.

\subsection{Mining Debian Packages}
\label{sec:dataset-generation}

\begin{figure*}
  \centering \usetikzlibrary{fit, positioning, shapes.symbols, decorations.pathreplacing}
\begin{tikzpicture}[
    debs/.style={draw, cloud, cloud puffs=11, cloud ignores aspect, align=center, font=\footnotesize},
    box/.style={draw, rectangle, align=center, font=\footnotesize, minimum width=24mm, minimum height=8mm}
]

\node (CORPUS GEN) [box] {Generation of\\corpus data};
\node (EMB CREATOR) [right=of CORPUS GEN, box] {Training of\\embedding model};
\node (PHASE 1) [draw, dashed, rectangle, fit=(CORPUS GEN) (EMB CREATOR), inner sep=4mm] {};
\node (TRAIN) [debs, left=of PHASE 1] {Training\\DEBs};
\node (PRE EMB) [right=of PHASE 1, align=center, font=\footnotesize] {Embedding\\Model};
\draw[decorate, decoration={brace, amplitude=1mm, raise=4mm}] (TRAIN.north east) -- (TRAIN.south east);

\node (RECORD CREATOR) [below=of PHASE 1, box] {Creation of\\dataset files};
\node (TEST) at (TRAIN|-RECORD CREATOR) [debs] {Test\\DEBs};
\node (RECORD) [right=of RECORD CREATOR, align=center, font=\footnotesize] {Labeled\\Dataset File};

\draw[-latex] ([xshift=4mm]TRAIN.east) -- (CORPUS GEN);
\draw[-latex] (CORPUS GEN) -- (EMB CREATOR);
\draw[-latex] (EMB CREATOR) -- (PRE EMB);

\draw[-latex] (TRAIN) -- (TRAIN|-PHASE 1.south) -- ++(0, -4mm) -- ++(2,0) |- ([yshift=1mm]RECORD CREATOR.west);
\draw[-latex] ([yshift=-1mm]TEST.east) -- ([yshift=-1mm]RECORD CREATOR.west);
\draw[-latex] (PRE EMB) -- (PRE EMB|-PHASE 1.south) -- ++(0,-4mm) -| (RECORD CREATOR);
\draw[-latex] (RECORD CREATOR) -- (RECORD);

\end{tikzpicture}
  \caption{Dataset creation pipeline. A function corpus is build from
    all available training packages. Based on this corpus the
    embeddings are trained. Using the pre-trained embeddings, each
    \debian package is transformed into a dataset file containing the
    embedded functions and corresponding labels.}
  \label{fig:dataset-generation}
\end{figure*}
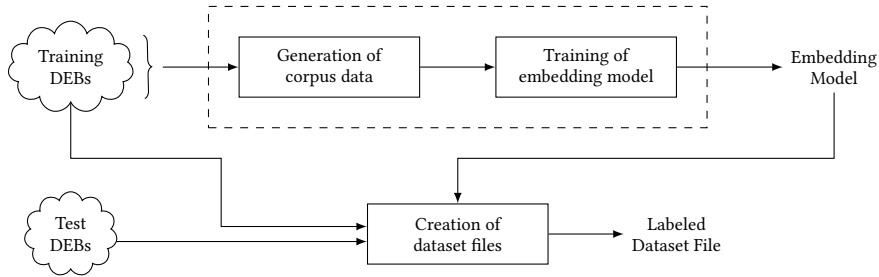

For the training of instruction embeddings with unlabeled data and the
automatic construction of a labeled dataset for training the
downstream tasks, we deploy two data processing pipelines. The first
pipeline processes \debian binary packages and emits the pre-trained
embeddings. The second pipeline uses \debian packages as well, but
also requires an embedding model to generate the labeled dataset used
to train and evaluate the downstream tasks. Both processes are
depicted in Figure~\ref{fig:dataset-generation}.

We manually build each package from source for the x86-64 architecture
using eight compilation combinations.  Each compilation combinations
uses a different compiler (\gcc or \clang) and one of four
optimization levels (\verb|O0| to \verb|O3|).  Note that building a
single \debian source package can produce multiple binary packages for
the same compilation combination.

In total, we extract \num{1293205} functions consisting of
\num{129487277} instructions from \num{480} binary \debian packages.
More details are provided in Table \ref{tab:dataset}, where we list
the number of binary packages (with and without variants due to
compilation combinations), functions, and instructions for each source
package. Since the dataset generation processes binary \debian
packages our dataset can be expanded by adding more packages. It is
also possible to create datasets for different architectures.

\begin{table}
  \caption{Our dataset is based on 480 binary \debian packages compiled
    from 8 source packages.}
  \centering \begin{tabular}{p{24mm}r@{\hskip 1mm}c@{\hskip 1mm}rrr}
\toprule
  \multicolumn{4}{r}{\#\,Packages} & \#\,Functions & \#\,Instructions \\
\addlinespace
  \multicolumn{6}{l}{Packages for training:} \\
  \midrule
  binutils & 208 & / & 26 & \num{1044289} & \num{110948051} \\
  coreutils & 8 & / & 1 & \num{101402} & \num{6585304} \\
  diffutils & 8 & / & 1 & \num{5889} & \num{529309} \\
  findutils & 16 & / & 2 & \num{9096} & \num{767734} \\
  \addlinespace
  Total & 240 & / & 30 & \num{1160676}  & \num{118830398} \\
  \addlinespace
  \multicolumn{6}{l}{Packages for testing:} \\
  \midrule
  inetutils & 88 & / & 11 & 19046 & 1684013 \\
  sg3-utils & 24 & / & 3 & 9412 & 1512412 \\
  usbutils & 8 & / & 1 & 1020 & 121158 \\
  util-linux & 120 & / & 15 & 103051 & 7339296 \\
  \addlinespace
  Total & 240 & / & 30 & \num{132529} & \num{10656879} \\
  \addlinespace
  \multicolumn{6}{l}{Full dataset:} \\
  \midrule
  Total & 480 & / & 60 & \num{1293205} & \num{129487277} \\
\bottomrule
\end{tabular}

  \label{tab:dataset}
\end{table}

\paragraph{Training and test splits} We split the dataset in training
and test data at the level of \debian source packages. Consequently,
all functions from a single source package will be placed either in
the training set or test set. We never spread functions from one
source package over training and test data. We believe that this is an
important detail as functions can be shared not only between the
binaries of a packages, but also between binary packages of a single
source package.  While functions may also be shared across source
packages, we argue that these cases are rare and do not necessarily
lead to overfitting. Instead, they reflect scenarios in which source
code is simply reused, for example, through the bad practice of
copying and pasting code snippets.
Our data split procedure is different from the approaches used by
\citeauthor{ChuSheSaxLia17} who divide each open source project into
training and test data or \citeauthor{PizIno21} who randomly partition
all functions of the whole dataset into training and test data. Our
procedure aims at reducing the risk that information from the training
data leak to the test data.

Table~\ref{tab:dataset} also shows which packages we use for training
and which we use for testing. The processing of training as well as
test packages are displayed in Figure~\ref{fig:dataset-generation}.

\paragraph{Unsupervised data} For pre-training embeddings, we create a
generic function corpus. To do so, we iterate over the binary \debian
packages selected for training and hand them over to the corpus
generator. The corpus generator creates a indefinitely repeating
stream of disassembled and formatted instruction sequences on-the-fly,
matching the requirements of the target embedding type. The streaming
approach allows for processing large amounts of data without storing
them in memory. Moreover, by working directly with \debian packages
the corpus can be easily extended.
This token stream is then used as unsupervised training data for the
\wtov, \itov, and \atov embeddings.
It is important to note that packages selected for testing are not
used for pre-training embeddings. This will prevent patterns from test
packages to leak into the training process.

In particular, we extract all ELF~binaries from each package.  Next,
we extract the opcodes of functions from the ELF~binaries.  Here, we
leverage the DWARF~\cite{manual:dwarf} debugging information to
identify every function and to pinpoint the address range of the
functions to extract the corresponding opcodes from the text segment
of the ELF binary. For this purpose, we use the python library
pyelftools\footnote{\url{https://github.com/eliben/pyelftools}} to
parse the text segment as well as the debugging information.  Last,
the opcodes are disassembled with the Capstone
engine\footnote{\url{https://www.capstone-engine.org/}} to produces a
sequence of tokens for each function. Since each embedding type
requires its specific instruction formatting and tokenization, we
implement different tokenization schemes for the \wtov, \itov, and
\atov embeddings.
Figure~\ref{fig:dataset-generation} shows where the corpus generator
fits into the pipeline.

\paragraph{Embedding Pre-training}

To pre-train the embedding models, we leverage the function corpus
extracted from Debian packages, as illustrated in
Figure~\ref{fig:dataset-generation}. In this process, we make use of
the gensim\footnote{\url{https://radimrehurek.com/gensim/}} Python
library, incorporating a module that implements the Word2Vec
algorithm. The Word2Vec algorithm is applied to both the \wtov and
\itov embedding types, as both are grounded in a Word2Vec model,
albeit with distinct encodings for each instruction. In the case of
\atov, we extend gensim to support the \atov embedding, as detailed in
Section~\ref{sec:instr-embedd} and
Section~\ref{sec:function-embeddings}. The training of instruction
embeddings marks the concluding step in the pipeline depicted in
Figure~\ref{fig:dataset-generation}. Subsequently, these resulting
embedding models are then reused to construct supervised datasets,
facilitating the training and evaluation of various tasks, as well as
the development of embedding models based on \Gemini and \SAFE.

\paragraph{Supervised data}

To create a labeled dataset for training the downstream tasks and
learning function embeddings we use the pre-trained embedding models
to encode the instruction sequences of functions and the debug
information to obtain the labels.  We create an independent record
file for each binary package and embedding. These dataset files
contain the encoded functions and the corresponding labels for all
four downstream tasks. Each file can then be loaded by
TensorFlow\footnote{\url{https://www.tensorflow.org/}} and fits
seamlessly into TensorFlow's data processing pipeline.

To train embedding models based on the network architectures of
\Gemini and \SAFE, labeled function pairs are required. While this
presents a limitation for \Gemini and \SAFE, both approaches can
incorporate pre-trained embeddings, allowing them to still benefit
from unsupervised data in theory. All embedding models undergo
training using a twin-network \cite{BroGuyLeC+93}.  The dataset used
for training the twin-networks is derived from the same Debian
packages used to pre-train the instruction embeddings and \atov.  This
involves a specialized corpus generator sampling both positive
function pairs (comprising the same functions with different
compilation configurations) and negative function pairs (consisting of
different functions from arbitrary packages).

All datasets used in our evaluation are available for other
researchers: \url{https://github.com/a0x77n/orbit-dataset}

\section{Experiments}
\label{sec:experiments}

Equipped with a large corpus of labeled code for different downstream
tasks, we are finally ready to compare the strengths and weaknesses of
the considered embeddings. To this end, we design five basic
experiments, each addressing a different aspect of pre-trained
embeddings.

\paragraph{Experimental instances.}

For every experiment, we evaluate multiple \emph{instances} of the
downstream tasks. In this context, an instance is defined by the
learning model of the task (T1-T4), the deployed embedding and the
available supervised training data. For example, the learning model
for task~T1 paired with a \wtov embedding and all training data from
Table~\ref{tab:dataset} is one instance of the compiler identification
task. The performance of each instance is then evaluated using the
test data from Table~\ref{tab:dataset}. With this experimental setup,
we can directly evaluate the capabilities of the embeddings, simply by
comparing the instances considered in one experiment.

\paragraph{Performance measures.}

As an evaluation criterion, we uee the \emph{accuracy} metric for
tasks T1-T4, aligning with its prominent usage as the primary measure
in the original publications~\cite{PizIno21, ChuSheSaxLia17}. Task T5
employs the \emph{area under the ROC curve}, a widely recognized
metric for assessing binary classifiers, providing a comprehensive
view of the trade-off between true-positive and false-positive
rates. The \emph{ROC curve} (receiver operating characteristic curve)
\citep{fawcett2006} visually represents the performance of a
classification model or detection system across various classification
thresholds, plotting the true positive rate against the false positive
rate.

Given a downstream task, the accuracy $\performance_F$ with respect to
a set of labeled test functions $F$ is given by the ratio of correct
classified functions $f \in F$ to the total number of functions in
$F$. The \emph{general accuracy} $\performance$ of an instance is the
performance over all test functions from Table~\ref{tab:dataset}.
Analogously, we define the accuracy of a binary package. For this
definition, we need a formal notion of a \emph{compilation
  combination} $c$, which is an element of
$C = \{ \text{\gcc}, \text{\clang} \} \times \{ \verb|O0|, \verb|O1|,
\verb|O2|, \verb|O3| \}$.  Now, given an instance of a downstream
task, the \emph{package accuracy} for the package~$p$, denoted as
$\performance_p$, is defined as the average accuracy over all
compilation combinations of $p$, that is,
$\frac{1}{8} \sum_{c \in C} \performance_{F(p_c)}$, where $F(p_c)$
denotes the functions in $p$ build with compilation combination $c$.%
We make use of these notions of general and package accuracy for the
different instances in our evaluation.

\paragraph{Experimental setup.}

Table~\ref{tab:dataset} gives an overview of the datasets
that are used for training and testing.
Each instance is evaluated on \num{132529}~functions from
\num{240}~binary \debian packages each.  The functions used for
training are determined by the instance, consistently sourced from
packages within the training set.  In total, \num{1160676}~functions
from \num{240}~packages are available for training. Notably, while we
vary the amount of supervised training data in some experiments, the
number of unsupervised training samples remains constant
throughout. This means that pre-trained embeddings are consistently
trained using the \emph{entire} set of training data.

Whenever we consider an instance involving \etoe learning, we add an
additional dropout layer after the embedding layer to counteract
overfitting. This layer randomly drops \qty{20}{\percent} of its
inputs.  The reason for this extension is that unlike pre-trained
embeddings, the size of the labeled data used to train instances with
\etoe learning varies, making them prone to overfitting when the
labeled data is limited.

We proceed with the presentation of the five distinct experiments,
each delineated in Sections~\ref{sec:experiments:baseline} to
\ref{sec:experiments:exp4}. Within each section, the purpose of the
experiment is elucidated alongside comprehensive details regarding its
execution. It is worth noting that the reader has the flexibility to
peruse these sections in any order they prefer. Furthermore, to
facilitate navigation and comprehension, each section closes with a
forward reference to the corresponding results expounded upon in
Section~\ref{sec:results}. Additionally, for convenient reference, an
overarching overview of all experiments is encapsulated in
Table~\ref{tab:experiments}, providing a succinct summary for quick
comprehension and navigation.

\begin{table}[htbp]
  \centering
  \caption{Overview of the experiments. The right columns list the
    corresponding sections for the description of the experiments and
    results.}
  \label{tab:experiments}
  \begin{tabular}{lrrrrr}
    \toprule
    Experiment & \#\,Shards & Dimension & Seq.~len. & Exp. & Res. \\
    \midrule
    Baseline & 1 & 128 (126\footnotemark) & 512 & \ref{sec:experiments:baseline} & \ref{sec:results:baseline} \\
    Experiment 1 & varies & 128 (126\footnotemark[5]) & 512 & \ref{sec:experiments:exp1} & \ref{sec:results:exp1} \\
    Experiment 2 & varies & 128 (126\footnotemark[5]) & 512 & \ref{sec:experiments:exp2} & \ref{sec:results:exp2} \\
    Experiment 3 & 1 & varies & 512 & \ref{sec:experiments:exp3} & \ref{sec:results:exp3} \\
    Experiment 4 & 1 & 512 & varies & \ref{sec:experiments:exp4} & \ref{sec:appendix:exp4} \\
    \bottomrule
  \end{tabular}
\end{table}
\footnotetext{The dimension of the \itov embedding is 126.}

\subsection{Baseline Experiment}
\label{sec:experiments:baseline}

To begin, we introduce a general baseline experiment. That is, we
design an experiment to test our implementations of the various
embeddings and downstream tasks for correctness.  We create five
different instances for each combination of downstream task and
embedding type and train them on all training data.  Each instance is
based on a different random seed in order to detect possible
variations. We further use this experiment to revise the peculiarities
of the downstream task and discuss the natural limits on the
classification performance and identify variations in the test data.
The results are presented in Section~\ref{sec:results:baseline} and
complemented in Appendix~\ref{sec:appendix:baseline}.

\subsection{Experiment 1: Size of Labeled Data}
\label{sec:experiments:exp1}

The objective of the first experiment is to investigate the effect of
the amount of labeled data on the accuracy of the employed
embeddings. Moreover, we consider the changes in training time.  The
rationale behind this experiment is the following: Unlike pre-trained
embeddings, an end-to-end embedding is build from scratch and can only
be trained with supervised data.

For each combination of downstream application and embedding type, we
create nine different groups of instances. The groups differ in the
amount of labeled data used for training. In contrast, the instances
within one group are trained with the same number of functions, but
the function sets are disjoint.
On that account, we partition the training data $T$ into $n$ disjoint
subsets $t_i^n, i=1, \dots, n$, i.e., $T = \bigcup_{i=1}^n t_i^n$.  We
refer to each set $t_i^n$ as a \emph{shard} (of the training data).
For instance, if $n = 1$ the same data that is used for pre-training
the embeddings is used to train the instances.  As $n$ gets larger
less supervised data is available for training the instances, e.g.,
for $n=2^8$ less than \qty{1}{\percent} of the data is used.
Doing so, we mimic scenarios where more unlabeled then labeled data is
available.
We conduct this experiment for $n = 2^i, i = 0, \dots 8$.  In total,
we train \num{511} ($\sum_{i=0}^8 2^i$) instances distributed over
nine instance groups for all \num{24} combinations of downstream task
and embedding type.

Note that the pre-trained embeddings remain unaffected by the value of
$n$.  They are consistently trained on the entire training set, while
only the learning model of the downstream task is trained on reduced
data, that is, it is trained on a single shard of the training data.
In contrast, experimental instances that use \etoe learning have only
this single shard to infer all weights for the embedding layer as well
as the learning model. Hence, the \etoe approaches face a significant
penalty for larger values of $n$.  Our primary findings of this
experiment are elaborated on in Section~\ref{sec:results:exp1}, while
Appendix~\ref{sec:appendix:exp1} offers supplementary insights.

\subsection{Experiment 2: Computational Expense}
\label{sec:experiments:exp2}

In the second experiment, we investigate the computational overhead of
\etoe learning, assuming pre-trained models are readily available and
thus only \etoe-learning incurs an additional run-time overhead. 
Using the setup from the previous section, we focus on the required
epochs, that is, the number of training cycles needed for each instance.

Recall that we work with groups of instances that exist for each task
and embedding pair, where  group is defined by the size of the
labeled data.  For example, one group uses the entire training data
while for another group it is split into multiple shards.  In this
case, the first group will contain a single instance and the
other group contains instances for each shard.  We compute the
\emph{mean number of epochs} for each group by averaging the number of
required epochs of its instances. We present the results in Section~\ref{sec:results:exp2}.

\subsection{Experiment 3: Embedding Dimension}
\label{sec:experiments:exp3}

The goal of the next experiment is to analyze the impact of the
embedding dimension on the general accuracy of an instance.  In case
of an end-to-end approach the embedding dimension is an easy to change
parameter of the network architecture.  It can be tuned like any other
hyper-parameter. By contrast, the dimension of a pre-trained embedding
is fixed.  If one wants to deploy a pre-trained embedding for some
application, the user needs to pick the embedding dimension in
advance.  In other cases, a particular pre-trained embedding might
only exist for a specific dimension. For that reason, it is worthwhile
to determine if it is possible to identify a recommended value for a
default embedding dimension that yields strong results across
different downstream tasks.

We create several pre-trained embeddings with
different embedding dimensions. We pick dimensions of the range from
\num{1} to \num{256}. Starting with the smallest possible dimension
for each embedding we subsequently increase the dimension in each
step. Some embeddings come with a restriction in regard to
the dimension: The embedding dimension of \itov is a multiple
of \num{9} and the smallest possible dimension of \atov is \num{2}.
\palmtree has a fixed dimensionality of \num{128}.
Again, all instances are trained on the entire dataset.
The results are summarized in Section~\ref{sec:results:exp2}.

\subsection{Experiment 4: Sequence Length}
\label{sec:experiments:exp4}

In the last experiment, we seek to find a viable sequence length for
the downstream tasks. We aim at determining a good
compromise between training time and the general accuracy of an
instance.  This trade-off enables us to train and evaluate thousands
of instances. To this end, we truncate each function sequence at
various positions and observe the general accuracy of the instances.
Starting with only the first instruction of each function, we double
the sequence length subsequently.  We stop at a maximum of \num{256}
instructions per function.  Figure~\ref{fig:function-length} shows
a bar plot of the distribution of function lengths.
Most functions are rather short. Only few functions are longer than \num{256} instructions. These functions are accumulated in the last bar of the
plots.
We use the \palmtree embedding for this experiment, as it is the only
context-dependent one.
We report the result in Section~\ref{sec:appendix:exp4} of the
appendix.

\section{Results and Discussion}
\label{sec:results}

We will now delve into the outcomes of the five experiments outlined
in Sections~\ref{sec:results:baseline} to \ref{sec:results:exp3}.  The
setup for each specific experiment can be found in
Section~\ref{sec:experiments}, where their objectives and designs are
described. Supplementary findings are also provided in the appendix
for further reference.

\subsection{Baseline Experiment}
\label{sec:results:baseline}

\begin{table}
  \caption{General accuracy of the baseline experiment. The second
    value (\textpm) reflects the minimal standard deviation observed for
    different random seeds, underscoring the stability of the results
    over all instances.}
  \centering 
%

\setlength{\tabcolsep}{4pt}
\begin{tabular}{lrrrrr}
\toprule
 & Task T1 & Task T2 & Task T3 & Task T4 & Average \\
  \addlinespace
  \multicolumn{6}{l}{Instruction embeddings:} \\
  \midrule
\wtov     & $0.97_{\pm .00}$ & $0.68_{\pm .00}$ & $0.80_{\pm .01}$ & $0.92_{\pm .05}$ & $0.84_{\pm .01}$ \\
\itovx    & $0.95_{\pm .00}$ & $0.67_{\pm .01}$ & $0.81_{\pm .01}$ & $0.92_{\pm .09}$ & $0.84_{\pm .01}$ \\
\atov     & $0.97_{\pm .00}$ & $0.68_{\pm .01}$ & $0.83_{\pm .00}$ & $0.93_{\pm .02}$ & $0.85_{\pm .00}$ \\
\palmtree & $0.97_{\pm .00}$ & $0.69_{\pm .01}$ & $0.86_{\pm .00}$ & $0.94_{\pm .07}$ & $0.86_{\pm .01}$ \\
\etoe     & $0.96_{\pm .00}$ & $0.68_{\pm .01}$ & $0.88_{\pm .01}$ & $0.94_{\pm .05}$ & $0.87_{\pm .01}$ \\
\rand     & $0.97_{\pm .00}$ & $0.67_{\pm .00}$ & $0.88_{\pm .01}$ & $0.94_{\pm .05}$ & $0.87_{\pm .00}$ \\
  \addlinespace
  \multicolumn{6}{l}{Function embeddings:} \\
  \midrule
\atov     & $0.93_{\pm .00}$ & $0.65_{\pm .00}$ & $0.43_{\pm .00}$ & $0.85_{\pm .03}$ & $0.72_{\pm .00}$ \\
\ada      & $0.87_{\pm .00}$ & $0.66_{\pm .00}$ & $0.60_{\pm .00}$ & $0.90_{\pm .07}$ & $0.76_{\pm .00}$ \\
\bottomrule
\end{tabular}
  \label{tab:baseline}
\end{table}

The outcome of the baseline experiment is summarized in
Table~\ref{tab:baseline} and part of
Figure~\ref{fig:limited-data-function-similarity}.
Table~\ref{tab:baseline} lists the general accuracy of experimental
instances that are trained on the entire dataset, that is, the same
amount of data is available to pre-training and \etoe learning. Our
results are in line with the original publications and show that our
re-implementations are correct.
\mbox{\cite{PizIno21, LiQuYin21}}.

In Figure~\ref{fig:limited-data-function-similarity}, it can be
observed that when utilizing \qty{100}{\percent} of the training data,
the area under the ROC curve ranges from \num{0.79} to \num{0.92} for
\Gemini and \num{0.79} to \num{0.90} for \SAFE. It is essential to
highlight that our dataset differs from the evaluation data previously
employed for function similarity \citep{DinFunCha19, MasAntPet+19,
  XuLiuFenYin+17} rendering direct comparisons challenging. Despite
these discrepancies, discernible patterns emerge: models using \wtov
and \palmtree demonstrate optimal performance, while the \etoe
approach yields comparable results to \atov. Conversely, \ada exhibits
the least favorable performance in this experiment.

Apart from the reproduced performance,
we observe \emph{no} notable differences between the embeddings.
This result is particularly striking for the \rand embedding that performs
as well as the specialized instruction embeddings.  We conclude that
the embedding type is not as important when sufficient labeled data is
available for the downstream task.
Furthermore, we observe that the results are stable across five runs
with different random seeds. This is quantified by the standard
deviation provided in Table~\ref{tab:baseline}.  The deviation is at
most \num{0.01}.  This suggests that the used optimization algorithms
are well calibrated and the networks capacities are adequate for the
data basis. As a result, the training algorithm is able to find model
weights that produce stable results for each downstream task.

\subsection{Experiment 1: Size of Labeled Data}
\label{sec:results:exp1}

\begin{figure*}
  \centering \begin{tikzpicture}[
    every mark/.append style={mark size=0.5pt}
]
    \begin{groupplot}[
        tiny,
        xmode=log, log basis x=2, x dir=reverse,
        xtick={1,2,4,8,16,32,64,128,256},
        xticklabels={100,50,25,\phantom{12.5},\phantom{6.25},,,,$<1$},
        group style={group size=5 by 1, xlabels at=edge bottom, ylabels at=edge left},
        xlabel=\% of training data,
        ylabel=Accuracy,
        legend to name=legend-limited-training-data, legend style={legend columns=6}
    ]

        \nextgroupplot[title=\titlestrut Compiler];

        \addplot+ coordinates {(1, 0.9698118902277992) (2, 0.9685087792105879) (4, 0.9692199443140747) (8, 0.9680824574244128) (16, 0.9652260071380604) (32, 0.9615619505919459) (64, 0.9565710618053407) (128, 0.950494915263829) (256, 0.9449924639135585)};
        \addplot+ coordinates {(1, 0.9475118653275887) (2, 0.9453591289453629) (4, 0.9444555531242219) (8, 0.940348904768013) (16, 0.9370651895056931) (32, 0.9264163409517917) (64, 0.9156465283070121) (128, nan) (256, 0.8265959419730776)};
        \addplot+ coordinates {(1, 0.9688943551977303) (2, 0.9682484588278791) (4, 0.9656301639641135) (8, 0.9645011657825834) (16, 0.9603629205683284) (32, 0.9558417402983498) (64, 0.9498801913166175) (128, 0.940228530170755) (256, 0.925065074200741)};
        \addplot+ coordinates {(1, 0.9721872194010368) (2, 0.9736397316813679) (4, 0.9734624120003924) (8, 0.9697585056855481) (16, 0.9686837409170823) (32, 0.9653019339163503) (64, 0.9595346820318572) (128, 0.9532779735944585) (256, 0.9406218697888764)};
        \addplot+ coordinates {(1, 0.9648605210934965) (2, 0.9644455175848304) (4, 0.958490217235473) (8, 0.9550060741422632) (16, 0.9460971749579338) (32, 0.9422949411072293) (64, 0.9318441624097368) (128, 0.9115397620520792) (256, 0.8939347708803356)};
        \addplot+ coordinates {(1, 0.9661704230772132) (2, 0.964072014427031) (4, 0.9628590723539754) (8, 0.9589882214458723) (16, 0.9559676561356383) (32, 0.947378262493492) (64, 0.9297441503369074) (128, 0.9018302715820689) (256, 0.8745529869877536)};
        \nextgroupplot[title=\titlestrut Optimization level]
        \addplot+ coordinates {(1, 0.6781142240566216) (2, 0.6823449961895133) (4, 0.6760671249311472) (8, 0.6717369405941341) (16, 0.6712177146134054) (32, 0.6678076119189008) (64, 0.6585766134204589) (128, 0.6541758077477382) (256, 0.6387944749356744)};
        \addplot+ coordinates {(1, 0.667639535497891) (2, 0.6696647526201812) (4, 0.6653845573421666) (8, 0.6630548785548823) (16, 0.6613179945521358) (32, 0.6513258418912087) (64, 0.6405320958054463) (128, 0.6270835062514619) (256, 0.6163874392585774)};
        \addplot+ coordinates {(1, 0.679196251386489) (2, 0.6784930090772585) (4, 0.6752578680892484) (8, 0.672147228153838) (16, 0.6672478287770979) (32, 0.6674102932188427) (64, 0.6589462259203646) (128, 0.6452810564102951) (256, 0.632044920357054)};
        \addplot+ coordinates {(1, 0.687564231224864) (2, 0.6854612952636782) (4, 0.6803284564133133) (8, 0.675701167291687) (16, 0.67532860732368) (32, 0.6715947547329264) (64, 0.6697722762565175) (128, 0.6624191096854274) (256, 0.651007544810947)};
        \addplot+ coordinates {(1, 0.6787601204264727) (2, 0.6802284782953165) (4, 0.671415312874918) (8, 0.6604469587788333) (16, 0.6691271344385002) (32, 0.6607065717691977) (64, 0.6379162768148858) (128, 0.6122424148676894) (256, 0.6102632784616951)};
        \addplot+ coordinates {(1, 0.6744833206317108) (2, 0.6729470530978126) (4, 0.6741335858566804) (8, 0.6732667944374439) (16, 0.6544289363082797) (32, 0.594046633189717) (64, 0.5824364384398887) (128, 0.5768024121135752) (256, 0.5797864877687148)};
        \nextgroupplot[ title=\titlestrut \#\,Parameters]
        \addplot+ coordinates {(1, 0.8044902499375723) (2, 0.7971328687203468) (4, 0.7981714376518126) (8, 0.784367362072749) (16, 0.7725789425892715) (32, 0.7573182220607931) (64, 0.7357764295929717) (128, 0.6889758580017101) (256, 0.5508977106554523)};
        \addplot+ coordinates {(1, 0.8136357101238716) (2, 0.8059030063638358) (4, 0.7831320514857778) (8, 0.7577183264852103) (16, 0.7292711856711539) (32, 0.7062933209991449) (64, 0.6535264759407656) (128, 0.5904193434882296) (256, 0.5338384119637845)};
        \addplot+ coordinates {(1, 0.8326409540456895) (2, 0.8300038591632426) (4, 0.8193571088057024) (8, 0.81380766989777) (16, 0.7917347695474185) (32, 0.7731580535439982) (64, 0.7409838170529613) (128, 0.6947442665887267) (256, 0.5861448531720809)};
        \addplot+ coordinates {(1, 0.8592313454859141) (2, 0.8582854721421382) (4, 0.8581568333673848) (8, 0.8432044675489774) (16, 0.8280950867555031) (32, 0.8149292392151521) (64, 0.7857642987673379) (128, 0.7380295590149296) (256, 0.5895430509523053)};
        \addplot+ coordinates {(1, 0.8763841910512815) (2, 0.8839489077054626) (4, 0.876274091394066) (8, 0.8633184642043692) (16, 0.8617298699234978) (32, 0.8547278060278617) (64, 0.8398003876188963) (128, 0.7944392579056094) (256, 0.5718207837222764)};
        \addplot+ coordinates {(1, 0.883963284980288) (2, 0.874641514002709) (4, 0.869338948037502) (8, 0.8555301430917195) (16, 0.8464776622551133) (32, 0.8356916320477022) (64, 0.81637287840609) (128, 0.7737835714096539) (256, 0.5952379474453853)};
        \nextgroupplot[title=\titlestrut Parameter type]
        \addplot+ coordinates {(1, 0.9166986656426994) (2, 0.9203745480144635) (4, 0.9131587789190746) (8, 0.9117238248376052) (16, 0.911466833061342) (32, 0.9082644355380628) (64, 0.9021530061038047) (128, 0.8921470762935586) (256, 0.8662876225460785)};
        \addplot+ coordinates {(1, 0.9150667178650284) (2, 0.9084549294422578) (4, 0.908838917154651) (8, 0.897113292374644) (16, 0.8926439353940674) (32, 0.8829327461521231) (64, 0.8622270337349205) (128, 0.8358966263079581) (256, 0.8223499035530863)};
        \addplot+ coordinates {(1, 0.9346628907874948) (2, 0.9303982272567278) (4, 0.9359920482544558) (8, 0.9353910674858404) (16, 0.9338526167162651) (32, 0.9311479532654955) (64, 0.9253312643995392) (128, 0.9201699920602541) (256, 0.9039271680806215)};
        \addplot+ coordinates {(1, 0.9403859076509552) (2, 0.9331141403475088) (4, 0.9304262263607564) (8, 0.9321801702345525) (16, 0.9309812086013247) (32, 0.9292117652235129) (64, 0.923630943809798) (128, 0.912706980876612) (256, 0.8728022578277496)};
        \addplot+ coordinates {(1, 0.9421810502063934) (2, 0.9369700169594573) (4, 0.9351500751975936) (8, 0.9338081181402195) (16, 0.9349055830213433) (32, 0.9362852888707561) (64, 0.942282221968897) (128, 0.9392665684698089) (256, 0.8832143308914114)};
        \addplot+ coordinates {(1, 0.9358052542318646) (2, 0.9291902659114908) (4, 0.9360220472944866) (8, 0.9345580941409874) (16, 0.9351245760135676) (32, 0.9360967949025631) (64, 0.936681651187162) (128, 0.9328497113092381) (256, 0.8892061391535471)};
        \nextgroupplot[title=\titlestrut Average]
        \addplot+ coordinates {(1, 0.8412335915283935) (2, 0.8409946094999187) (4, 0.8381174420497305) (8, 0.8328930612870178) (16, 0.8289916509483641) (32, 0.8225678719492929) (64, 0.8120442789368364) (128, 0.7951467995701005) (256, 0.7487144228789866)};
        \addplot+ coordinates {(1, 0.8348396733146309) (2, 0.8312677728521778) (4, 0.8242816655975029) (8, 0.8134101854600116) (16, 0.8038673008243731) (32, 0.7904895493235545) (64, 0.7667074085865018) (128, nan) (256, 0.6981463947440231)};
        \addplot+ coordinates {(1, 0.8526993671189069) (2, 0.8506687858449106) (4, 0.8478280018766217) (8, 0.8452038432446406) (16, 0.8369561626916117) (32, 0.8305015056059286) (64, 0.8173060904033779) (128, 0.7984517272579301) (256, 0.7598738890471357)};
        \addplot+ coordinates {(1, 0.8637576477121492) (2, 0.8616125542161753) (4, 0.8595889624005438) (8, 0.8541106153595741) (16, 0.8496326943883267) (32, 0.8440715270530337) (64, 0.8334289397087407) (128, 0.8152802246201277) (256, 0.7620438151671246)};
        \addplot+ coordinates {(1, 0.8644343804754746) (2, 0.8653686699922446) (4, 0.8592428405924762) (8, 0.8519752400831075) (16, 0.8517778836303055) (32, 0.8472342569631283) (64, 0.8364561890217631) (128, 0.8125866501345231) (256, 0.7378627950772192)};
        \addplot+ coordinates {(1, 0.8640732265446224) (2, 0.859208372030677) (4, 0.8594951313156458) (8, 0.8544478806621794) (16, 0.8467453539251075) (32, 0.826744731576075) (64, 0.8145741875951475) (128, 0.7943653027249313) (256, 0.7325698437185834)};

        \legend{\wtov, \itov, \atov, \palmtree, \etoe, \rand}

    \end{groupplot}
    \node[below] at (current bounding box.south) {\pgfplotslegendfromname{legend-limited-training-data}};
\end{tikzpicture}
  \caption{General accuracy of the considered embeddings in relation
    to the available labeled training data (shards)}
  \label{fig:limited-data}
\end{figure*}

\begin{figure}
  \centering \begin{tikzpicture}[
    every mark/.append style={mark size=0.5pt}
]
    \begin{groupplot}[
        tiny,width=32mm,height=22mm,
        xmode=log, log basis x=2, x dir=reverse,
        group style={group size=2 by 1, xlabels at=edge bottom, ylabels at=edge left},
        xlabel=\% of training data,
        ylabel={Area under the ROC curve},
        scale only axis,
        legend pos=outer north east, 
        title={Function similarity},
        enlarge y limits,
        xtick={1,2,4,8,16,32,64,128,256},
        ytick={0.7,0.75,0.8,0.85,0.9,0.95},
        xticklabels={100,50,25,\phantom{12.5},\phantom{6.25},,,,$<1$},
        legend to name=legend-limited-training-data-similarity, legend style={legend columns=4}
    ]

    \nextgroupplot[ymin=0.75,ymax=0.92, smooth,title={\titlestrut \Gemini}];
    
    \addplot coordinates {(1,0.9102343784005628) (2,0.9111686280769193) (4,0.9122626935574701) (8,0.9153214744969032) (16,0.909977642526845) (32,0.8963661386928927) (64,0.8840940383229919) (128,0.8677788181677157) (256,0.7956931787531792)}; \addlegendentry{\Gemini w/~\wtov};
    \addplot coordinates {(1,0.8170037102676525) (2,0.8167114120651611) (4,0.8264385699212713) (8,0.8229559804857377) (16,0.8244497671073883) (32,0.8197442213721499) (64,0.818895680106315) (128,0.8143556592596751) (256,0.8063299877255534)}; \addlegendentry{\Gemini w/~\itovx};
    \addplot coordinates {(1,0.88266491877696) (2,0.8773679783393373) (4,0.8906834450714046) (8,0.8807675955641358) (16,0.8765792334490208) (32,0.8613421257240168) (64,0.8574280466988293) (128,0.8340201887304844) (256,0.8000719086800148)}; \addlegendentry{\Gemini w/~\atov};
    \addplot coordinates {(1,0.910740543623465) (2,0.9175993153968046) (4,0.9225703604498837) (8,0.917738840664589) (16,0.9090406832503956) (32,0.9058446504485055) (64,0.8930387777177361) (128,0.8703618898391272) (256,0.8519480735513094)}; \addlegendentry{\Gemini w/~\palmtree};
    \addplot coordinates {(1,0.8827759771785425) (2,0.8833876524779888) (4,0.8805249525204495) (8,0.8683761362252422) (16,0.850289375534703) (32,0.8441001593374842) (64,0.843100862118408) (128,0.8172343848522042) (256,0.7732827884911516)}; \addlegendentry{\Gemini w/~\etoe};
    \addplot coordinates {(1,0.8205068189419833) (2,0.8345422145165646) (4,0.8455795163974573) (8,0.8327771085306562) (16,0.8450825448123764) (32,0.841217025970291) (64,0.8438380758384397) (128,0.853793265301371) (256,0.8406637793859432)}; \addlegendentry{\Gemini w/~\rand};

    \addplot+[opacity=0.0, mark=otimes*, mark options={draw=black,fill=red,opacity=1.0}] coordinates {(1,0.79)}; \addlegendentry{\ada};
    \addplot+[opacity=0.0, mark=otimes*, mark options={draw=black,fill=blue,opacity=1.0}] coordinates {(1,0.88)}; \addlegendentry{\atov};

    \addplot+[mark=*] coordinates {(1,0.9156169981251984) (2,0.922161832497211) (4,0.9224697156426205) (8,0.9186053808748261) (16,0.9117744572282027) (32,0.9134176479305363) (64,0.8876820092897423) (128,0.8847098815167765) (256,0.859532657604306)}; \addlegendentry{\Gemini};

    \nextgroupplot[ymin=0.7,ymax=0.9,smooth,title={\titlestrut \SAFE}];

    \addplot coordinates {(1,0.8825642704646453) (2,0.8918608377732573) (4,0.8829866007550854) (8,0.8853656889783821) (16,0.8777663516666964) (32,0.8660917339351728) (64,0.8299358841267196) (128,0.7950371197031574) (256,0.7823636857510298)};  \addlegendentry{\SAFE};
    \addplot coordinates {(1,0.8753143400303315) (2,0.8868574551077467) (4,0.8888900116136276) (8,0.8910636640871705) (16,0.8776820409243802) (32,0.8878065873058008) (64,0.8474192114009191) (128,0.8377414215994183) (256,0.7751647096535588)}; \addlegendentry{\SAFE w/~\itovx};
    \addplot coordinates {(1,0.8922595576167704) (2,0.893922485946981) (4,0.8887115082500657) (8,0.8838556804850147) (16,0.8786686849070414) (32,0.8516260799514191) (64,0.8282442089065281) (128,0.8055878167019177) (256,0.785739088344311)}; \addlegendentry{\SAFE w/~\atov};
    \addplot coordinates {(1,0.9040725691694356) (2,0.8912982157852555) (4,0.8888852415303014) (8,0.8899131072802884) (16,0.8814887771174003) (32,0.8614973784622127) (64,0.8417101664369274) (128,0.8355322206611816) (256,0.7678682549570496)}; \addlegendentry{\SAFE w/~\palmtree};
    \addplot coordinates {(1,0.8900603144431629) (2,0.8849858935142874) (4,0.8834008823334497) (8,0.8731837920091615) (16,0.8700915996936602) (32,0.8603854613503937) (64,0.8374759957884685) (128,0.8300002459090057) (256,0.8098264730515077)}; \addlegendentry{\SAFE w/~\etoe};
    \addplot coordinates {(1,0.8668842458395769) (2,0.8398296028312165) (4,0.853587606431251) (8,0.8532307746231007) (16,0.8428529846460595) (32,0.8315132558040192) (64,0.8197299297087295) (128,0.7959136285512444) (256,0.7935058296133906)}; \addlegendentry{\SAFE w/~\rand};

    \addplot+[opacity=0.0, mark=otimes*, mark options={draw=black,fill=red,opacity=1.0}] coordinates {(1,0.79)}; \addlegendentry{\ada};
    \addplot+[opacity=0.0, mark=otimes*, mark options={draw=black,fill=blue,opacity=1.0}] coordinates {(1,0.88)}; \addlegendentry{\atov};

    \legend{w/ \wtov, w/ \itov, w/ \atov, w/ \palmtree, w/ \etoe, w/ \rand, \ada, \atov, \Gemini};

    \end{groupplot}
    \node[below] at (current bounding box.south) {\pgfplotslegendfromname{legend-limited-training-data-similarity}};

\end{tikzpicture}
  \caption{General accuracy of the considered embeddings in relation to the available labeled training data (shards).\vspace*{-0.35cm}}
  \label{fig:limited-data-function-similarity}
\end{figure}

The preceding results suggest that the amount of labeled data
available is a key factor in the success of \etoe learning. Therefore,
we turn to the first experiment in which this quantity is varied. The
corresponding results for tasks~T1-T4 are shown in
Figure~\ref{fig:limited-data}, and
Figure~\ref{fig:limited-data-function-similarity} shows the results
for task~T5.

The plots show the performance of the embeddings trained with
different amounts of labeled data. This amount is defined by the
number of shards used to partition the training data. For example, if
8 shards have been generated from the training data, only
$\frac{1}{8}$ or \qty{12.5}{\percent} of the labeled data is available
for supervised training of the learning models or, in the case of
task~T5, the embedding models.

Expectedly, the overall performance improves with fewer shards; that
is, the more supervised training data is available, the better the
prediction.  This trend can be observed for all embedding types and
downstream tasks. At a certain point, however, this improvement
saturates, so that increasing the amount of labeled data does not
improve the performance. We conclude that the size of our evaluation
corpus is sufficient for the downstream tasks and further extensions
of it would not change the results significantly.

When we investigate the left part of the plots, that is, instances
with limited labeled data, we observe different yet subtle performance
differences. In the case of tasks~T1 and T2, the pre-trained
embeddings \wtov, \atov, and \palmtree lead to better results when
less labeled data is used. The \etoe approach requires more labeled
data to attain a comparable performance. This confirms the expectation
one has for pre-trained embeddings: Since a vector representation for
the input is already learned, the neural network has fewer degrees of
freedom and thus can better generalize even if only a few labeled
examples are given.
This observation is also confirmed by the performance of the \rand
embedding. Despite its simplicity, it also provides reasonable
accuracy, suggesting that it is not the embedding that matters, but
rather the reduced number of trainable weights of the networks.

Our findings for tasks~T3 and T4 are the other way around. For both
tasks, the pre-trained embeddings \wtov, \atov, and \palmtree show no
benefit compared to the \etoe approach.  Even with less data
available, the \etoe approach gives better results. For both tasks,
pre-training negatively affects the performance, and we cannot
identify a general advantage over conventional learning. This is
further contradicted by the results of the \rand embedding, which
works just as well, even when trained on limited training data.

The function embeddings employed in task T5 exhibit a slightly
different trend: when utilizing minimal data, no discernible
differences can be observed. However, with an increased volume of
available data, \Gemini models with \wtov and \palmtree, as well as
the manual selected features of the original \Gemini implementation,
perform better than the \etoe-based models and \atov, which have a
similar performance. Such differences cannot be identified for \SAFE
embedding models. In this context, the \etoe embedding type and all
pre-trained embeddings exhibit similar results, whereas the
performance of the \rand embedding is comparatively weaker.
Surprisingly, the performance of \itov shifts from being subpar in the
context of \Gemini to displaying a remarkably competent performance
with the \SAFE model. In both scenarios, the \ada embedding fails to
achieve competitive results.
Notably, in this task, the \rand embedding finally reveals its
limitations. In the case of \Gemini, this embedding type fails to
derive any benefit from a larger pool of supervised training data.

We must conclude that for the considered downstream tasks,
pre-training is not necessarily beneficial. When we examine the
average performance on the right-hand side of
Figure~\ref{fig:limited-data}, \etoe learning and \palmtree provide
the best results, regardless of the amount of labeled data
available. These results suggest that when developing a learning-based
method for binary code analysis, experimenting with \etoe learning is
clearly a sound and likely successful design decision.

\paragraph{Minimizing the labeled data}
In addition, Table~\ref{tab:accuracy_drop} shows the general
performance averaged over all instances that are trained with
256~shards.  In this case, each shard contains only
approximately~\num{4500} labeled training samples. Further, the table
shows the margin to the baseline experiment with
approximately~\num{1160676} labeled samples. Considering task~T2, the
accuracy of the \etoe approach drops by \num{0.08} while the
pre-trained approaches drop by \num{0.04} (\wtov and \palmtree),
\num{0.6} (\itov), and \num{0.05} (\atov) only. Similar observations
can be made for task~T1. As already observed, the tasks~T3 and T4
behave differently, i.e., all embedding types lose accuracy just about
equally. The limitation of labeled data has the most impact on task~T3
with a loss of up to \num{0.30}. Task~T4, however, is barely affected,
which is attracting our attention. We show in the appendix that a
different metric gives a different picture in this case.

\begin{table}[tbp]
  \caption{General accuracy of the considered embeddings with a
    minimum of labeled data. The second value (\textdownarrow)
    displayed represents the margin to the accuracy on the
    full dataset.}
  \centering 
%

\setlength{\tabcolsep}{4pt}
\begin{tabular}{lrrrrr}
\toprule
 & Task T1 & Task T2 & Task T3 & Task T4 & Average \\
  \addlinespace
  \multicolumn{6}{l}{Instruction embeddings:} \\
  \midrule
\wtov     & $0.94_{\downarrow .02}$ & $0.64_{\downarrow .04}$ & $0.55_{\downarrow .25}$ & $0.87_{\downarrow .05}$ & $0.75_{\downarrow .09}$ \\
\itovx    & $0.83_{\downarrow .12}$ & $0.62_{\downarrow .06}$ & $0.53_{\downarrow .30}$ & $0.82_{\downarrow .09}$ & $0.70_{\downarrow .14}$ \\
\atov     & $0.93_{\downarrow .04}$ & $0.63_{\downarrow .05}$ & $0.59_{\downarrow .25}$ & $0.90_{\downarrow .02}$ & $0.76_{\downarrow .09}$ \\
\palmtree & $0.94_{\downarrow .03}$ & $0.65_{\downarrow .04}$ & $0.59_{\downarrow .26}$ & $0.87_{\downarrow .07}$ & $0.76_{\downarrow .10}$ \\
\etoe     & $0.89_{\downarrow .07}$ & $0.61_{\downarrow .08}$ & $0.57_{\downarrow .29}$ & $0.88_{\downarrow .05}$ & $0.74_{\downarrow .12}$ \\
\rand     & $0.87_{\downarrow .09}$ & $0.58_{\downarrow .10}$ & $0.60_{\downarrow .30}$ & $0.89_{\downarrow .05}$ & $0.73_{\downarrow .13}$ \\
  \addlinespace
  \multicolumn{6}{l}{Function embeddings:} \\
  \midrule
\atov     & $0.86_{\downarrow .07}$ & $0.62_{\downarrow .03}$ & $0.39_{\downarrow .04}$ & $0.82_{\downarrow .03}$ & $0.67_{\downarrow .04}$ \\
\ada      & $0.71_{\downarrow .06}$ & $0.50_{\downarrow .16}$ & $0.33_{\downarrow .27}$ & $0.83_{\downarrow .07}$ & $0.59_{\downarrow .16}$ \\
\bottomrule
\end{tabular}
  \label{tab:accuracy_drop}
\end{table}

\subsection{Experiment 2: Computational Expense}
\label{sec:results:exp2}

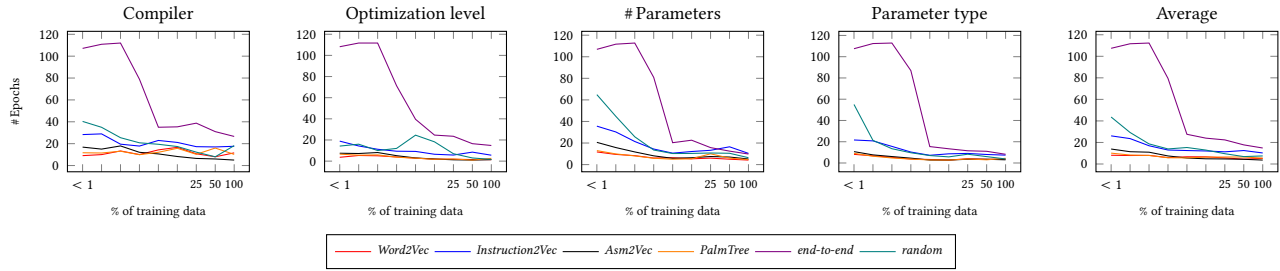
\begin{figure*}
  \centering
  \begin{tikzpicture}[
    every mark/.append style={mark size=0.5pt}
]
    \begin{groupplot}[
        tiny,
        xtick={1,2,4,8,16,32,64,128,256},
        xticklabels={100,50,25,\phantom{12.5},\phantom{6.25},,,,$<1$},
        xmode=log, log basis x=2, x dir=reverse,
        group style={group size=5 by 1, xlabels at=edge bottom, ylabels at=edge left},
        xlabel=\% of training data,
        ylabel=\#\,Epochs,
    ]

        \nextgroupplot[legend to name=legend-epochs-chart, legend style={legend columns=6}, title=\mystrut Compiler]

        \addplot+ coordinates {(1, 11.6) (2, 8.0) (4, 10.75) (8, 16.625) (16, 14.3125) (32, 10.03125) (64, 13.328125) (128, 9.9921875) (256, 9.04296875)};
        \addplot+ coordinates {(1, 17.6) (2, 17.0) (4, 17.25) (8, 21.125) (16, 22.9375) (32, 17.8125) (64, 19.53125) (128, 28.9453125) (256, 28.30078125)};
        \addplot+ coordinates {(1, 5.0) (2, 6.0) (4, 6.5) (8, 8.25) (16, 10.625) (32, 12.09375) (64, 17.921875) (128, 15.0234375) (256, 16.84375)};
        \addplot+ coordinates {(1, 10.2) (2, 16.0) (4, 10.25) (8, 15.75) (16, 12.125) (32, 9.90625) (64, 13.09375) (128, 11.453125) (256, 11.69140625)};
        \addplot+ coordinates {(1, 26.6) (2, 31.0) (4, 38.75) (8, 35.375) (16, 35.0) (32, 78.90625) (64, 112.0) (128, 110.8671875) (256, 107.109375)};
        \addplot+ coordinates {(1, 18.2) (2, 8.0) (4, 12.25) (8, 17.375) (16, 19.4375) (32, 20.78125) (64, 25.453125) (128, 34.96875) (256, 40.34375)};

        \addlegendentry{\wtov};
        \addlegendentry{\itov};
        \addlegendentry{\atov};
        \addlegendentry{\palmtree};
        \addlegendentry{\etoe};
        \addlegendentry{\rand};

        \nextgroupplot[legend to name=legend-epochs-chart, legend style={legend columns=6}, title=\mystrut Optimization level]

        \addplot+ coordinates {(1, 1.4) (2, 1.5) (4, 1.75) (8, 1.875) (16, 2.8125) (32, 3.9375) (64, 5.046875) (128, 5.328125) (256, 3.62109375)};
        \addplot+ coordinates {(1, 5.6) (2, 8.5) (4, 5.75) (8, 6.5) (16, 9.1875) (32, 9.375) (64, 11.21875) (128, 14.3359375) (256, 18.90234375)};
        \addplot+ coordinates {(1, 1.4) (2, 1.0) (4, 1.25) (8, 2.0) (16, 3.0) (32, 5.15625) (64, 8.125) (128, 7.203125) (256, 7.46484375)};
        \addplot+ coordinates {(1, 2.6) (2, 1.0) (4, 2.0) (8, 1.75) (16, 2.8125) (32, 3.71875) (64, 5.890625) (128, 5.609375) (256, 6.59375)};
        \addplot+ coordinates {(1, 14.8) (2, 16.5) (4, 23.5) (8, 24.625) (16, 39.5625) (32, 71.28125) (64, 111.671875) (128, 111.6796875) (256, 108.1875)};
        \addplot+ coordinates {(1, 1.6) (2, 3.0) (4, 7.0) (8, 18.25) (16, 24.5625) (32, 12.03125) (64, 10.0) (128, 16.015625) (256, 14.2265625)};

        \addlegendentry{\wtov};
        \addlegendentry{\itov};
        \addlegendentry{\atov};
        \addlegendentry{\palmtree};
        \addlegendentry{\etoe};
        \addlegendentry{\rand};

        \nextgroupplot[legend to name=legend-epochs-chart, legend style={legend columns=6}, title=\mystrut \#\,Parameters]

        \addplot+ coordinates {(1, 4.2) (2, 5.0) (4, 6.0) (8, 5.375) (16, 6.375) (32, 6.0625) (64, 8.296875) (128, 9.4453125) (256, 11.73046875)};
        \addplot+ coordinates {(1, 10.4) (2, 16.5) (4, 13.25) (8, 12.125) (16, 10.625) (32, 14.4375) (64, 21.421875) (128, 30.28125) (256, 35.65625)};
        \addplot+ coordinates {(1, 5.2) (2, 7.0) (4, 7.5) (8, 6.375) (16, 5.8125) (32, 8.125) (64, 11.75) (128, 15.796875) (256, 20.625)};
        \addplot+ coordinates {(1, 4.6) (2, 6.0) (4, 10.0) (8, 5.375) (16, 5.0) (32, 5.875) (64, 8.046875) (128, 9.90625) (256, 12.96484375)};
        \addplot+ coordinates {(1, 9.8) (2, 12.5) (4, 15.75) (8, 22.625) (16, 20.3125) (32, 80.875) (64, 112.71875) (128, 111.671875) (256, 106.87109375)};
        \addplot+ coordinates {(1, 6.4) (2, 10.5) (4, 10.75) (8, 10.375) (16, 10.25) (32, 13.5) (64, 25.78125) (128, 44.7734375) (256, 64.87109375)};

        \addlegendentry{\wtov};
        \addlegendentry{\itov};
        \addlegendentry{\atov};
        \addlegendentry{\palmtree};
        \addlegendentry{\etoe};
        \addlegendentry{\rand};

        \nextgroupplot[legend to name=legend-epochs-chart, legend style={legend columns=6}, title=\mystrut Parameter type]

        \addplot+ coordinates {(1, 3.6) (2, 3.5) (4, 4.0) (8, 2.75) (16, 3.3125) (32, 3.71875) (64, 5.34375) (128, 6.7734375) (256, 8.1171875)};
        \addplot+ coordinates {(1, 7.4) (2, 8.0) (4, 9.0) (8, 8.625) (16, 7.25) (32, 10.3125) (64, 15.625) (128, 20.796875) (256, 21.6484375)};
        \addplot+ coordinates {(1, 3.0) (2, 3.5) (4, 3.5) (8, 2.75) (16, 2.75) (32, 4.5625) (64, 6.0625) (128, 7.6328125) (256, 10.6875)};
        \addplot+ coordinates {(1, 4.4) (2, 3.0) (4, 3.75) (8, 2.75) (16, 2.875) (32, 3.65625) (64, 4.734375) (128, 6.375) (256, 9.07421875)};
        \addplot+ coordinates {(1, 8.2) (2, 11.0) (4, 11.5) (8, 13.375) (16, 15.4375) (32, 86.9375) (64, 112.84375) (128, 112.359375) (256, 107.51171875)};
        \addplot+ coordinates {(1, 3.8) (2, 6.0) (4, 8.25) (8, 5.75) (16, 7.0) (32, 9.625) (64, 13.4375) (128, 21.28125) (256, 55.0546875)};

        \addlegendentry{\wtov};
        \addlegendentry{\itov};
        \addlegendentry{\atov};
        \addlegendentry{\palmtree};
        \addlegendentry{\etoe};
        \addlegendentry{\rand};

        \nextgroupplot[legend to name=legend-epochs-chart, legend style={legend columns=6}, title=\mystrut Average]

        \addplot+ coordinates {(1, 5.2) (2, 4.5) (4, 5.625) (8, 6.65625) (16, 6.703125) (32, 5.9375) (64, 8.00390625) (128, 7.884765625) (256, 8.1279296875)};
        \addplot+ coordinates {(1, 10.25) (2, 12.5) (4, 11.3125) (8, 12.09375) (16, 12.5) (32, 12.984375) (64, 16.94921875) (128, 23.58984375) (256, 26.126953125)};
        \addplot+ coordinates {(1, 3.65) (2, 4.375) (4, 4.6875) (8, 4.84375) (16, 5.546875) (32, 7.484375) (64, 10.96484375) (128, 11.4140625) (256, 13.9052734375)};
        \addplot+ coordinates {(1, 5.45) (2, 6.5) (4, 6.5) (8, 6.40625) (16, 5.703125) (32, 5.7890625) (64, 7.94140625) (128, 8.3359375) (256, 10.0810546875)};
        \addplot+ coordinates {(1, 14.85) (2, 17.75) (4, 22.375) (8, 24.0) (16, 27.578125) (32, 79.5) (64, 112.30859375) (128, 111.64453125) (256, 107.419921875)};
        \addplot+ coordinates {(1, 7.5) (2, 6.875) (4, 9.5625) (8, 12.9375) (16, 15.3125) (32, 13.984375) (64, 18.66796875) (128, 29.259765625) (256, 43.6240234375)};

        \addlegendentry{\wtov};
        \addlegendentry{\itov};
        \addlegendentry{\atov};
        \addlegendentry{\palmtree};
        \addlegendentry{\etoe};
        \addlegendentry{\rand};

    \end{groupplot}
    \node[below] at (current bounding box.south) {\pgfplotslegendfromname{legend-epochs-chart}};
\end{tikzpicture}
  \caption{Training time of the considered embeddings in relation to
    the supervised data size.}
  \label{fig:train-time}
\end{figure*}

\begin{figure}
  \centering


    




\begin{tikzpicture}[
    every mark/.append style={mark size=0.5pt}
]
    \begin{groupplot}[
        tiny,width=32mm,height=22mm,
        xmode=log, log basis x=2, x dir=reverse,
        group style={group size=2 by 1, xlabels at=edge bottom, ylabels at=edge left},
        xlabel=\% of training data,
        ylabel={\# Epochs},
        scale only axis,
        legend pos=outer north east, 
        title={Function similarity},
        enlarge y limits,
        xtick={1,2,4,8,16,32,64,128,256},
        xticklabels={100,50,25,\phantom{12.5},\phantom{6.25},,,,$<1$},
        legend to name=legend-epochs-chart-function-similarity, legend style={legend columns=4}
    ]

    \nextgroupplot[title={\titlestrut \Gemini}];

    \addplot coordinates { (1,16.0) (2,16.0) (4,17.75) (8,13.0) (16,12.75) (32,11.5) (64,11.125) (128,11.0) (256,11.5) };
    \addplot coordinates { (1,37.0) (2,23.5) (4,20.25) (8,17.0) (16,18.625) (32,16.125) (64,14.625) (128,15.75) (256,14.625) };
    \addplot coordinates { (1,26.0) (2,20.0) (4,18.75) (8,13.5) (16,12.125) (32,11.25) (64,11.625) (128,11.0) (256,11.0) };
    \addplot coordinates { (1,18.0) (2,15.5) (4,18.75) (8,12.75) (16,11.75) (32,11.5) (64,11.125) (128,11.25) (256,11.0) };
    \addplot coordinates { (1,33.0) (2,30.5) (4,25.25) (8,18.0) (16,18.625) (32,14.875) (64,12.875) (128,11.625) (256,11.375) };
    \addplot coordinates { (1,25.0) (2,20.5) (4,20.75) (8,15.875) (16,18.875) (32,19.5) (64,18.625) (128,15.25) (256,13.0) };
    \addplot+[draw=black, mark=*] coordinates { (1,30.0) (2,30.0) (4,21.75) (8,15.0) (16,15.875) (32,13.875) (64,13.125) (128,11.875) (256,11.625) };

    \nextgroupplot[title={\titlestrut \SAFE}];

    \addplot coordinates { (1,19.0) (2,26.5) (4,19.5) (8,14.5) (16,14.625) (32,13.875) (64,12.375) (128,13.25) (256,12.25) };
    \addplot coordinates { (1,24.0) (2,20.0) (4,17.5) (8,15.625) (16,14.0) (32,14.75) (64,14.0) (128,12.375) (256,15.25) };
    \addplot coordinates { (1,26.0) (2,22.5) (4,19.5) (8,14.75) (16,12.375) (32,13.75) (64,12.0) (128,11.5) (256,11.125) };
    \addplot coordinates { (1,27.0) (2,18.5) (4,15.25) (8,13.375) (16,12.1) (32,11.636363636363637) (64,11.0) (128,11.125) (256,12.0) };
    \addplot coordinates { (1,36.0) (2,20.5) (4,15.25) (8,18.625) (16,16.5) (32,13.25) (64,13.375) (128,13.0) (256,11.875) };
    \addplot coordinates { (1,57.0) (2,30.5) (4,23.5) (8,22.5) (16,16.875) (32,24.0) (64,23.0) (128,16.5) (256,14.5) };

    \legend{w/~\wtov, w/~\itov, w/~\atov, w/~\palmtree, w/~\etoe, w/~\rand, \Gemini};

    \end{groupplot}
    \node[below] at (current bounding box.south) {\pgfplotslegendfromname{legend-epochs-chart-function-similarity}};

\end{tikzpicture}
  \caption{Number of epochs needed to train embeddings
    for code similarity in relation to the supervised data size.}
  \label{fig:train-time-function-similarity}
  \vspace{-0.15cm}
\end{figure}
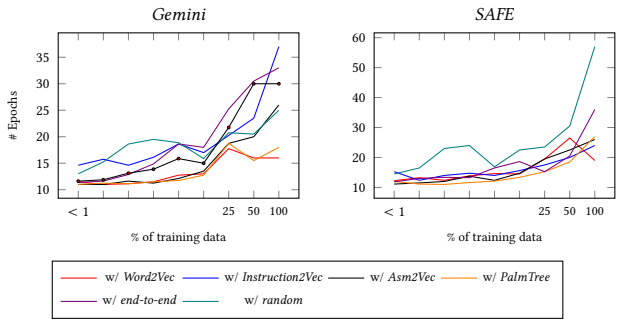

While we've noted advantages of \etoe learning in binary code analysis,
they do come with a prize: Figure~\ref{fig:train-time} shows the training time for
tasks T1-T4.
It stands out that the \etoe approach needs
the most epochs for training.
When training on more labeled data, we notice a significant drop in the
number of epochs.
Especially for tasks~T3 and T4,
where the \etoe learner has a comparable training time once the full
dataset is used.
The epochs needed for instances with pre-trained embeddings
remain relatively constant.
We conclude that pre-trained embeddings have acquired relevant
knowledge about the instructions, which helps training instances for
each downstream application with fewer epochs.

An alternative perspective is presented for task~T5 in
Figure~\ref{fig:train-time-function-similarity}. Here, we maintain
a consistent number of training samples per epoch.
regardless of the available data. Consequently, training the embedding
models for \Gemini and \SAFE with a larger volume of supervised data
necessitates more epochs to cycle through all samples. Our
observations indicate a general trend: the training algorithm
effectively reduces the loss over more epochs when more data is
available and concludes earlier with less data, yielding less
favorable results, as depicted in
Figure~\ref{fig:limited-data-function-similarity}.
Among the different embedding models, those based on \palmtree
demonstrate the fastest training time while delivering one of the best
results. Notably, the \Gemini model with the \itov embedding trains
slowly, contrasting with its relatively short training time for
\SAFE. Unlike tasks T1-T4, the training time of \etoe-based models is
not as significantly impacted negatively; however, it is still
noticeably slower.

We can conclude here that learning-based methods for code analysis can
profit from a pre-trained embedding as the training process becomes
faster. However, this is only relevant in scenarios where a complex
network architecture is involved and the training time per epoch is
high or the hardware resources are limited. Note that the prediction
time is not affected, which is much more relevant in practice. We want
to stress that we do not address the training time of the pre-trained
embeddings, since at least in theory those embeddings are only trained
once, but used for many
tasks.

\subsection{Experiment 3: Embedding Dimension}
\label{sec:results:exp3}

It remains to investigate the role of the embedding dimension in our
analysis, focusing specifically on tasks T1-T4 and the instruction
embeddings.  The results of this experiment are depicted in
Figure~\ref{fig:dimension-chart} in the appendix. The plot shows the
shift of the general accuracy of instances trained with different
dimensions.  Recall that the lowest dimension of \atov is \num{2} and
\itov's embedding dimension is a multiple of \num{9}. The \palmtree
embedding, that we use as an off-the-shelf embedding, has a fixed
dimension of \num{128}.

This experiment shows that the \etoe approach adapts better to lower
embedding dimensions. On average, instances with an \etoe embedding
attain the best performance already with 16 dimensions.  The
pre-trained embedding types tend to require higher dimensionality. As
a general rule, a dimension of \num{128} suffices for our tasks. This
confirms the observation made by \citeauthor{LiQuYin21}

In summary, we have seen that an \etoe embedding can get good results
out of lower embedding dimensions. All pre-trained embedding types
require a higher dimensionality to yield equal results. Our
explanation is that pre-trained embeddings are more universal while an
\etoe embedding is always tailored towards the downstream task. Hence,
an \etoe embedding does not require the same complexity as pre-trained
embeddings.

\subsection{Discussion}
\label{sec:discussion}

Our experiments shed new light on the role of pre-trained embeddings
in binary code analysis.
We show that an \etoe approach can compete with pre-trained embeddings
if sufficient labeled data is available. If training time is not a
constraint, \etoe learning can obtain the same performance even with
smaller embedding dimension.  For pre-trained embeddings to be
beneficial, the labeled data must be orders of magnitude smaller than
the unlabeled data. This raises the question of whether these
embeddings are actually relevant in practice, since the learning model
of the downstream task requires labeled data in any case.

Moreover, our evaluation reveals that the embedding process is not
crucial for the downstream task, as the success of the \rand embedding
striking illustrates. We credit this finding to task-agnostic
relations among instructions that are not as advanced as the
connections between words in natural language. Without a given
analysis task, an instruction embedding cannot carry the same amount
of information as a natural word embedding. This makes instruction
embeddings less useful and also explains why the differences between
our pre-trained embeddings are negligible.

Overall, our empirical analysis demonstrates that a general benefit of
pre-training does not exist for binary code analysis in practice. This
is contradictory to previous research in our domain
\citep[e.g.,][]{LiQuYin21}, and the common strategy to favor
pre-trained embeddings over \etoe learning. Interestingly, research in
other fields also arrived at this observation \cite{WanKhaMa20} and
questioned the role of pre-training if sufficient labeled information
is available.
Thus, our analysis refutes the intuition that pre-training is a
generally beneficial and therefore mandatory step in designing methods
for binary analysis.

Based on our observations, we work out the following recommendations
for deep learning in binary code analysis.

\vspace{0.2cm}
\begin{enumerate}
\setlength{\itemsep}{4pt}

\item[R1] Our first recommendation is to always try \etoe learning, as
  it produces tailored embeddings for the task at hand and is easy to
  implement. Consequently, \etoe learning should always be considered
  as the baseline approach. 

\item[R2] When there is little labeled data or the size of the
  unlabeled data is orders of magnitude larger, the use of a
  pre-trained embedding is indicated. We recommend the use of the
  \palmtree model as it performs best in our evaluation.

\item[R3] We also recommend to differentiate between training and
  inference time. Only if the training time for a model is
  constrained, pre-trained embeddings are clearly preferable over slow
  \etoe learning.

\item[R4] Our last recommendation is to always publish pre-trained
  embeddings. If researchers need to re-train them anyway, their
  benefit vanishes and again \etoe learning becomes an viable
  alternative.

\end{enumerate}

Our recommendations are to be understood as guidelines and must be
adapted to the respective downstream task. Nevertheless, we argue that
pre-training is not a ``silver bullet'' in our field and conventional
learning concepts must not be neglected from the start when developing
new approaches for binary code analysis.

\section{Conclusion}
\label{sec:conclusion}

This paper investigates the role of pre-training on binary code
analysis. To this end, we compare four pre-trained embeddings, an
end-to-end approach, and an random instruction embedding under
different downstream tasks. Our results show that binary analysis
tasks with sufficiently labeled data do not benefit from pre-trained
embeddings. Instead, conventional \etoe learning provides the best
performance on average. Only if the labeled data is artificially
reduced, we can observe an advantage of pre-training.

Our results have consequences for applying deep learning in other
tasks of binary code analysis. First, an \etoe setup is typically easy
to deploy along with the learning model. Hence, if labeled data is
available, it should be the first option when developing a new
approach. Second, when labeled data is scarce or the training time is
constrained, pre-trained embeddings are a reasonable solution. For
example, we could show that \palmtree often provides good results when
we capped the available labeled data.

Overall, our study highlights an interesting aspect of
interdisciplinary research. The benefits of embeddings in one research
area can only be transferred to another if the experimental setup is
limited by similar constraints. Once these constraints are lifted from
a setup, the performance improvement may disappear. Therefore, we
recommend that practitioners develop learning-based approaches
judiciously and, when in doubt, prefer well-known learning concepts
over the latest inventions.

\section*{Acknowledgements}
This work was funded by Deutsche Forschungsgemeinschaft (DFG, German Research Foundation) under Germany’s Excellence Strategy -- EXC 2092 CASA -- (390781972), the European Research Council (ERC) under the consolidator grant MALFOY (101043410), and the German Federal Ministry of Education and Research under the grant BIFOLD24B.
\clearpage

\bibliographystyle{ACM-Reference-Format}
\bibliography{sec, orbit}


\begin{thebibliography}{39}


\ifx \showCODEN    \undefined \def \showCODEN     #1{\unskip}     \fi
\ifx \showDOI      \undefined \def \showDOI       #1{#1}\fi
\ifx \showISBNx    \undefined \def \showISBNx     #1{\unskip}     \fi
\ifx \showISBNxiii \undefined \def \showISBNxiii  #1{\unskip}     \fi
\ifx \showISSN     \undefined \def \showISSN      #1{\unskip}     \fi
\ifx \showLCCN     \undefined \def \showLCCN      #1{\unskip}     \fi
\ifx \shownote     \undefined \def \shownote      #1{#1}          \fi
\ifx \showarticletitle \undefined \def \showarticletitle #1{#1}   \fi
\ifx \showURL      \undefined \def \showURL       {\relax}        \fi
\providecommand\bibfield[2]{#2}
\providecommand\bibinfo[2]{#2}
\providecommand\natexlab[1]{#1}
\providecommand\showeprint[2][]{arXiv:#2}

\bibitem[Aghajanyan et~al\mbox{.}(2021)]%
        {AghGupShr21}
\bibfield{author}{\bibinfo{person}{Armen Aghajanyan}, \bibinfo{person}{Anchit Gupta}, \bibinfo{person}{Akshat Shrivastava}, \bibinfo{person}{Xilun Chen}, \bibinfo{person}{Luke Zettlemoyer}, {and} \bibinfo{person}{Sonal Gupta}.} \bibinfo{year}{2021}\natexlab{}.
\newblock \showarticletitle{Muppet: Massive Multi-task Representations with Pre-Finetuning}. In \bibinfo{booktitle}{\emph{Proc. of the Conference on Empirical Methods in Natural Language Processing}}. \bibinfo{pages}{5799--5811}.
\newblock


\bibitem[Ahn et~al\mbox{.}(2022)]%
        {AhnAhnKooPae+22}
\bibfield{author}{\bibinfo{person}{Sunwoo Ahn}, \bibinfo{person}{Seonggwan Ahn}, \bibinfo{person}{Hyungjoon Koo}, {and} \bibinfo{person}{Yunheung Paek}.} \bibinfo{year}{2022}\natexlab{}.
\newblock \showarticletitle{Practical Binary Code Similarity Detection with BERT-based Transferable Similarity Learning.}. In \bibinfo{booktitle}{\emph{Proc. of the Annual Computer Security Applications Conference ({ACSAC})}}. \bibinfo{pages}{361--374}.
\newblock


\bibitem[Alves-Foss and Song(2019)]%
        {AlvSon19}
\bibfield{author}{\bibinfo{person}{Jim Alves-Foss} {and} \bibinfo{person}{Jia Song}.} \bibinfo{year}{2019}\natexlab{}.
\newblock \showarticletitle{Function boundary detection in stripped binaries}. In \bibinfo{booktitle}{\emph{Proc. of the Annual Computer Security Applications Conference ({ACSAC})}}. \bibinfo{pages}{84--96}.
\newblock


\bibitem[Biswas et~al\mbox{.}(2022)]%
        {BisBarLaz+22}
\bibfield{author}{\bibinfo{person}{Sajib Biswas}, \bibinfo{person}{Timothy Barao}, \bibinfo{person}{John Lazzari}, \bibinfo{person}{Jeret McCoy}, \bibinfo{person}{Xiuwen Liu}, {and} \bibinfo{person}{Alexander Kostandarithes}.} \bibinfo{year}{2022}\natexlab{}.
\newblock \showarticletitle{Geometric Analysis and Metric Learning of Instruction Embeddings}. In \bibinfo{booktitle}{\emph{Proc. of the International Joint Conference on Neural Networks ({IJCNN})}}.
\newblock


\bibitem[Bromley et~al\mbox{.}(1993)]%
        {BroGuyLeC+93}
\bibfield{author}{\bibinfo{person}{Jane Bromley}, \bibinfo{person}{Isabelle Guyon}, \bibinfo{person}{Yann LeCun}, \bibinfo{person}{Eduard S\"{a}ckinger}, {and} \bibinfo{person}{Roopak Shah}.} \bibinfo{year}{1993}\natexlab{}.
\newblock \showarticletitle{Signature Verification using a "Siamese" Time Delay Neural Network}. In \bibinfo{booktitle}{\emph{Advances in Neural Information Processing Systems}}, Vol.~\bibinfo{volume}{6}.
\newblock


\bibitem[Chen et~al\mbox{.}(2018)]%
        {CheShiLi+18}
\bibfield{author}{\bibinfo{person}{Yu Chen}, \bibinfo{person}{Zhiqiang Shi}, \bibinfo{person}{Hong Li}, \bibinfo{person}{Weiwei Zhao}, \bibinfo{person}{Yiliang Liu}, {and} \bibinfo{person}{Yuansong Qiao}.} \bibinfo{year}{2018}\natexlab{}.
\newblock \showarticletitle{{HIMALIA}: Recovering Compiler Optimization Levels from Binaries by Deep Learning}. In \bibinfo{booktitle}{\emph{Proc. of the Intelligent Systems Conference ({IntelliSys})}}.
\newblock


\bibitem[Chua et~al\mbox{.}(2017a)]%
        {ChuSheSaxLia+17}
\bibfield{author}{\bibinfo{person}{Zheng~Leong Chua}, \bibinfo{person}{Shiqi Shen}, \bibinfo{person}{Prateek Saxena}, {and} \bibinfo{person}{Zhenkai Liang}.} \bibinfo{year}{2017}\natexlab{a}.
\newblock \showarticletitle{Neural Nets Can Learn Function Type Signatures From Binaries.}. In \bibinfo{booktitle}{\emph{Proc. of the {USENIX} Security Symposium}}. \bibinfo{pages}{99--116}.
\newblock


\bibitem[Chua et~al\mbox{.}(2017b)]%
        {ChuSheSaxLia17}
\bibfield{author}{\bibinfo{person}{Zheng~Leong Chua}, \bibinfo{person}{Shiqi Shen}, \bibinfo{person}{Prateek Saxena}, {and} \bibinfo{person}{Zhenkai Liang}.} \bibinfo{year}{2017}\natexlab{b}.
\newblock \showarticletitle{Neural Nets Can Learn Function Type Signatures From Binaries}. In \bibinfo{booktitle}{\emph{Proc. of the {USENIX} Security Symposium}}. \bibinfo{pages}{99--116}.
\newblock


\bibitem[Dai et~al\mbox{.}(2016)]%
        {DaiDaiSong16}
\bibfield{author}{\bibinfo{person}{Hanjun Dai}, \bibinfo{person}{Bo Dai}, {and} \bibinfo{person}{Le Song}.} \bibinfo{year}{2016}\natexlab{}.
\newblock \showarticletitle{Discriminative Embeddings of Latent Variable Models for Structured Data}. In \bibinfo{booktitle}{\emph{Proc. of the International Conference on Machine Learning ({ICML})}}. \bibinfo{pages}{2702--2711}.
\newblock


\bibitem[Devlin et~al\mbox{.}(2019)]%
        {DevChaLeeTou19}
\bibfield{author}{\bibinfo{person}{Jacob Devlin}, \bibinfo{person}{Ming{-}Wei Chang}, \bibinfo{person}{Kenton Lee}, {and} \bibinfo{person}{Kristina Toutanova}.} \bibinfo{year}{2019}\natexlab{}.
\newblock \showarticletitle{{BERT:} Pre-training of Deep Bidirectional Transformers for Language Understanding}. In \bibinfo{booktitle}{\emph{Proc. of the Conference of The North American Chapter of the Association for Computational Linguistics: Human Language Technologies ({NAACL-HLT})}}.
\newblock


\bibitem[Ding et~al\mbox{.}(2019)]%
        {DinFunCha19}
\bibfield{author}{\bibinfo{person}{Steven Ding}, \bibinfo{person}{Benjamin Fung}, {and} \bibinfo{person}{Philippe Charland}.} \bibinfo{year}{2019}\natexlab{}.
\newblock \showarticletitle{Asm2Vec: Boosting Static Representation Robustness for Binary Clone Search against Code Obfuscation and Compiler Optimization}. In \bibinfo{booktitle}{\emph{Proc. of the {IEEE} Symposium on Security and Privacy}}.
\newblock


\bibitem[Donahue et~al\mbox{.}(2014)]%
        {DonJiaVin14}
\bibfield{author}{\bibinfo{person}{Jeff Donahue}, \bibinfo{person}{Yangqing Jia}, \bibinfo{person}{Oriol Vinyals}, \bibinfo{person}{Judy Hoffman}, \bibinfo{person}{Ning Zhang}, \bibinfo{person}{Eric Tzeng}, {and} \bibinfo{person}{Trevor Darrell}.} \bibinfo{year}{2014}\natexlab{}.
\newblock \showarticletitle{DeCAF: A Deep Convolutional Activation Feature for Generic Visual Recognition}. In \bibinfo{booktitle}{\emph{Proc. of the 31st International Conference on Machine Learning}}. \bibinfo{pages}{647--655}.
\newblock


\bibitem[DWARF Debugging Information Format Committee(2010)]%
        {manual:dwarf}
DWARF Debugging Information Format Committee \bibinfo{year}{2010}\natexlab{}.
\newblock \bibinfo{booktitle}{\emph{{DWARF} debugging information format}}.
\newblock DWARF Debugging Information Format Committee.
\newblock
\newblock
\shownote{Version~4}.


\bibitem[Edunov et~al\mbox{.}(2018)]%
        {EduOttAul18}
\bibfield{author}{\bibinfo{person}{Sergey Edunov}, \bibinfo{person}{Myle Ott}, \bibinfo{person}{Michael Auli}, {and} \bibinfo{person}{David Grangier}.} \bibinfo{year}{2018}\natexlab{}.
\newblock \showarticletitle{Understanding Back-Translation at Scale}. In \bibinfo{booktitle}{\emph{Proc. of the Conference on Empirical Methods in Natural Language Processing}}. \bibinfo{pages}{489--50}.
\newblock


\bibitem[Fawcett(2006)]%
        {fawcett2006}
\bibfield{author}{\bibinfo{person}{Tom Fawcett}.} \bibinfo{year}{2006}\natexlab{}.
\newblock \showarticletitle{An introduction to ROC analysis}.
\newblock \bibinfo{journal}{\emph{Pattern recognition letters}} \bibinfo{volume}{27}, \bibinfo{number}{8} (\bibinfo{year}{2006}), \bibinfo{pages}{861--874}.
\newblock


\bibitem[Feng et~al\mbox{.}(2016)]%
        {FengZhou+16}
\bibfield{author}{\bibinfo{person}{Qian Feng}, \bibinfo{person}{Rundong Zhou}, \bibinfo{person}{Chengcheng Xu}, \bibinfo{person}{Yao Cheng}, \bibinfo{person}{Brian Testa}, {and} \bibinfo{person}{Heng Yin}.} \bibinfo{year}{2016}\natexlab{}.
\newblock \showarticletitle{Scalable Graph-based Bug Search for Firmware Images}. In \bibinfo{booktitle}{\emph{Proc. of the {ACM} Conference on Computer and Communications Security ({CCS})}}. \bibinfo{pages}{480--491}.
\newblock


\bibitem[Guo et~al\mbox{.}(2019)]%
        {GuoMuXin19}
\bibfield{author}{\bibinfo{person}{Wenbo Guo}, \bibinfo{person}{Dongliang Mu}, \bibinfo{person}{Xinyu Xing}, \bibinfo{person}{Min Du}, {and} \bibinfo{person}{Dawn Song}.} \bibinfo{year}{2019}\natexlab{}.
\newblock \showarticletitle{{DEEPVSA}: Facilitating Value-set Analysis with Deep Learning for Postmortem Program Analysis}. In \bibinfo{booktitle}{\emph{Proc. of the {USENIX} Security Symposium}}.
\newblock


\bibitem[He et~al\mbox{.}(2018)]%
        {HeIvaTsaRay+18}
\bibfield{author}{\bibinfo{person}{Jingxuan He}, \bibinfo{person}{Pesho Ivanov}, \bibinfo{person}{Petar Tsankov}, \bibinfo{person}{Veselin Raychev}, {and} \bibinfo{person}{Martin~T. Vechev}.} \bibinfo{year}{2018}\natexlab{}.
\newblock \showarticletitle{Debin: Predicting Debug Information in Stripped Binaries.}. In \bibinfo{booktitle}{\emph{Proc. of the {ACM} Conference on Computer and Communications Security ({CCS})}}. \bibinfo{pages}{1667--1680}.
\newblock


\bibitem[Jin et~al\mbox{.}(2022)]%
        {JinPeiWonLin+22}
\bibfield{author}{\bibinfo{person}{Xin Jin}, \bibinfo{person}{Kexin Pei}, \bibinfo{person}{Jun~Yeon Won}, {and} \bibinfo{person}{Zhiqiang Lin}.} \bibinfo{year}{2022}\natexlab{}.
\newblock \showarticletitle{SymLM: Predicting Function Names in Stripped Binaries via Context-Sensitive Execution-Aware Code Embeddings.}. In \bibinfo{booktitle}{\emph{Proc. of the {ACM} Conference on Computer and Communications Security ({CCS})}}. \bibinfo{pages}{1631--1645}.
\newblock


\bibitem[Le and Mikolov(2014)]%
        {LeMik14}
\bibfield{author}{\bibinfo{person}{Quoc Le} {and} \bibinfo{person}{Tomas Mikolov}.} \bibinfo{year}{2014}\natexlab{}.
\newblock \showarticletitle{Distributed Representations of Sentences and Documents}. In \bibinfo{booktitle}{\emph{Proc. of the International Conference on Machine Learning ({ICML})}}.
\newblock


\bibitem[Lee et~al\mbox{.}(2019)]%
        {LeeKwoChoLimBaePar19}
\bibfield{author}{\bibinfo{person}{Yongjun Lee}, \bibinfo{person}{Hyun Kwon}, \bibinfo{person}{Sang-Hoon Choi}, \bibinfo{person}{Seung-Ho Lim}, \bibinfo{person}{Sung~Hoon Baek}, {and} \bibinfo{person}{Ki-Woong Park}.} \bibinfo{year}{2019}\natexlab{}.
\newblock \showarticletitle{Instruction2vec: Efficient Preprocessor of Assembly Code to Detect Software Weakness with CNN}.
\newblock \bibinfo{journal}{\emph{Applied Sciences}}  \bibinfo{volume}{9} (\bibinfo{year}{2019}).
\newblock


\bibitem[Lee et~al\mbox{.}(2017)]%
        {LeeCho+17}
\bibfield{author}{\bibinfo{person}{Young~Jun Lee}, \bibinfo{person}{Sang-Hoon Choi}, \bibinfo{person}{Chulwoo Kim}, \bibinfo{person}{Seung-Ho Lim}, {and} \bibinfo{person}{Ki-Woong Park}.} \bibinfo{year}{2017}\natexlab{}.
\newblock \showarticletitle{Learning Binary Code with Deep Learning to Detect Software Weakness}. In \bibinfo{booktitle}{\emph{Proc. of the International Conference on Internet ({ICONI})}}.
\newblock


\bibitem[Li et~al\mbox{.}(2021)]%
        {LiQuYin21}
\bibfield{author}{\bibinfo{person}{Xuezixiang Li}, \bibinfo{person}{Yu Qu}, {and} \bibinfo{person}{Heng Yin}.} \bibinfo{year}{2021}\natexlab{}.
\newblock \showarticletitle{PalmTree: Learning an Assembly Language Model for Instruction Embedding}. In \bibinfo{booktitle}{\emph{Proc. of the {ACM} Conference on Computer and Communications Security ({CCS})}}.
\newblock


\bibitem[Lin et~al\mbox{.}(2017)]%
        {LinCheCoh17}
\bibfield{author}{\bibinfo{person}{Di Lin}, \bibinfo{person}{Guangyong Chen}, \bibinfo{person}{Daniel Cohen-Or}, \bibinfo{person}{Pheng-Ann Heng}, {and} \bibinfo{person}{Hui Huang}.} \bibinfo{year}{2017}\natexlab{}.
\newblock \showarticletitle{Cascaded feature network for semantic segmentation of RGB-D images}. In \bibinfo{booktitle}{\emph{Proc. of the {IEEE} international conference on computer vision}}. \bibinfo{pages}{1311--1319}.
\newblock


\bibitem[Massarelli et~al\mbox{.}(2019a)]%
        {MasLunPet+19}
\bibfield{author}{\bibinfo{person}{Luca Massarelli}, \bibinfo{person}{Giuseppe~Antonio Di~Luna}, \bibinfo{person}{Fabio Petroni}, \bibinfo{person}{Roberto Baldoni}, {and} \bibinfo{person}{Leonardo Querzoni}.} \bibinfo{year}{2019}\natexlab{a}.
\newblock \showarticletitle{SAFE: Self-Attentive Function Embeddings for Binary Similarity}. In \bibinfo{booktitle}{\emph{Proc. of the Conference on Detection of Intrusions and Malware {\&} Vulnerability Assessment ({DIMVA})}}.
\newblock


\bibitem[Massarelli et~al\mbox{.}(2019b)]%
        {MasAntPet+19}
\bibfield{author}{\bibinfo{person}{Luca Massarelli}, \bibinfo{person}{Giuseppe Antonio~Di Luna}, \bibinfo{person}{Fabio Petroni}, \bibinfo{person}{Roberto Baldoni}, {and} \bibinfo{person}{Leonardo Querzoni}.} \bibinfo{year}{2019}\natexlab{b}.
\newblock \showarticletitle{{SAFE:} Self-Attentive Function Embeddings for Binary Similarity}. In \bibinfo{booktitle}{\emph{Proc. of the Conference on Detection of Intrusions and Malware {\&} Vulnerability Assessment ({DIMVA})}}.
\newblock


\bibitem[Mikolov et~al\mbox{.}(2013a)]%
        {MikCheCorDea13}
\bibfield{author}{\bibinfo{person}{Tomas Mikolov}, \bibinfo{person}{Kai Chen}, \bibinfo{person}{Greg Corrado}, {and} \bibinfo{person}{Jeffrey Dean}.} \bibinfo{year}{2013}\natexlab{a}.
\newblock \showarticletitle{Efficient Estimation of Word Representations in Vector Space}. In \bibinfo{booktitle}{\emph{Proc. of the International Conference on Learning Representations ({ICLR Workshop})}}.
\newblock


\bibitem[Mikolov et~al\mbox{.}(2013b)]%
        {MikSutCheCorDea13}
\bibfield{author}{\bibinfo{person}{Tomas Mikolov}, \bibinfo{person}{Ilya Sutskever}, \bibinfo{person}{Kai Chen}, \bibinfo{person}{Greg Corrado}, {and} \bibinfo{person}{Jeff Dean}.} \bibinfo{year}{2013}\natexlab{b}.
\newblock \showarticletitle{Distributed representations of words and phrases and their compositionality}.
\newblock \bibinfo{journal}{\emph{Advances in Neural Information Processing Systems}}  \bibinfo{volume}{26} (\bibinfo{year}{2013}).
\newblock


\bibitem[Pei et~al\mbox{.}(2021a)]%
        {PeiGuaBroChe+21}
\bibfield{author}{\bibinfo{person}{Kexin Pei}, \bibinfo{person}{Jonas Guan}, \bibinfo{person}{Matthew Broughton}, \bibinfo{person}{Zhongtian Chen}, \bibinfo{person}{Songchen Yao}, \bibinfo{person}{David Williams-King}, \bibinfo{person}{Vikas Ummadisetty}, \bibinfo{person}{Junfeng Yang}, \bibinfo{person}{Baishakhi Ray}, {and} \bibinfo{person}{Suman Jana}.} \bibinfo{year}{2021}\natexlab{a}.
\newblock \showarticletitle{StateFormer: {F}ine-Grained Type Recovery from Binaries using Generative State Modeling}. In \bibinfo{booktitle}{\emph{Proc. of the {ACM} Joint European Software Engineering Conference and Symposium on the Foundations of Software Engineering}}.
\newblock


\bibitem[Pei et~al\mbox{.}(2021b)]%
        {PeiGuaWil+21}
\bibfield{author}{\bibinfo{person}{Kexin Pei}, \bibinfo{person}{Jonas Guan}, \bibinfo{person}{David Williams-King}, \bibinfo{person}{Junfeng Yang}, {and} \bibinfo{person}{Suman Jana}.} \bibinfo{year}{2021}\natexlab{b}.
\newblock \showarticletitle{{XDA}: Accurate, Robust Disassembly with Transfer Learning}. In \bibinfo{booktitle}{\emph{Proc. of the Network and Distributed System Security Symposium ({NDSS})}}.
\newblock


\bibitem[Pizzolotto and Inoue(2021)]%
        {PizIno21}
\bibfield{author}{\bibinfo{person}{Davide Pizzolotto} {and} \bibinfo{person}{Katsuro Inoue}.} \bibinfo{year}{2021}\natexlab{}.
\newblock \showarticletitle{Identifying Compiler and Optimization Level in Binary Code From Multiple Architectures}.
\newblock \bibinfo{journal}{\emph{IEEE Access}}  \bibinfo{volume}{9} (\bibinfo{year}{2021}).
\newblock


\bibitem[Redmond et~al\mbox{.}(2019)]%
        {RedLuoZen19}
\bibfield{author}{\bibinfo{person}{Kimberly Redmond}, \bibinfo{person}{Lannan Luo}, {and} \bibinfo{person}{Qiang Zeng}.} \bibinfo{year}{2019}\natexlab{}.
\newblock \showarticletitle{A cross-architecture instruction embedding model for natural language processing-inspired binary code analysis}. In \bibinfo{booktitle}{\emph{Proc. of the Workshop on Binary Analysis Research ({BAR})}}.
\newblock


\bibitem[Shin et~al\mbox{.}(2015)]%
        {ShiSonMoa15}
\bibfield{author}{\bibinfo{person}{Eui Chul~Richard Shin}, \bibinfo{person}{Dawn Song}, {and} \bibinfo{person}{Reza Moazzezi}.} \bibinfo{year}{2015}\natexlab{}.
\newblock \showarticletitle{Recognizing Functions in Binaries with Neural Networks.}. In \bibinfo{booktitle}{\emph{Proc. of the {USENIX} Security Symposium}}. \bibinfo{pages}{611--626}.
\newblock


\bibitem[Vaswani et~al\mbox{.}(2017)]%
        {ShaParUsz+17}
\bibfield{author}{\bibinfo{person}{Ashish Vaswani}, \bibinfo{person}{Noam Shazeer}, \bibinfo{person}{Niki Parmar}, \bibinfo{person}{Jakob Uszkoreit}, \bibinfo{person}{Llion Jones}, \bibinfo{person}{Aidan Gomez}, \bibinfo{person}{Lukasz Kaiser}, {and} \bibinfo{person}{Illia Polosukhin}.} \bibinfo{year}{2017}\natexlab{}.
\newblock \showarticletitle{Attention is All you Need}. In \bibinfo{booktitle}{\emph{Advances in Neural Information Processing Systems}}.
\newblock


\bibitem[Wang et~al\mbox{.}(2020)]%
        {WanKhaMa20}
\bibfield{author}{\bibinfo{person}{Sinong Wang}, \bibinfo{person}{Madian Khabsa}, {and} \bibinfo{person}{Hao Ma}.} \bibinfo{year}{2020}\natexlab{}.
\newblock \showarticletitle{To Pretrain or Not to Pretrain: Examining the Benefits of Pretrainng on Resource Rich Tasks}. In \bibinfo{booktitle}{\emph{Proc. of Annual Meeting of the Association for Computational Linguistics ({ACL})}}.
\newblock


\bibitem[Xu et~al\mbox{.}(2017a)]%
        {XuLiuFeng+17}
\bibfield{author}{\bibinfo{person}{Xiaojun Xu}, \bibinfo{person}{Chang Liu}, \bibinfo{person}{Qian Feng}, \bibinfo{person}{Heng Yin}, \bibinfo{person}{Le Song}, {and} \bibinfo{person}{Song Dawn}.} \bibinfo{year}{2017}\natexlab{a}.
\newblock \showarticletitle{Neural Network-based Graph Embedding for Cross-Platform Binary Code Similarity Detection}. In \bibinfo{booktitle}{\emph{Proc. of the {ACM} Conference on Computer and Communications Security ({CCS})}}.
\newblock


\bibitem[Xu et~al\mbox{.}(2017b)]%
        {XuLiuFenYin+17}
\bibfield{author}{\bibinfo{person}{Xiaojun Xu}, \bibinfo{person}{Chang Liu}, \bibinfo{person}{Qian Feng}, \bibinfo{person}{Heng Yin}, \bibinfo{person}{Le Song}, {and} \bibinfo{person}{Dawn Song}.} \bibinfo{year}{2017}\natexlab{b}.
\newblock \showarticletitle{Neural Network-based Graph Embedding for Cross-Platform Binary Code Similarity Detection.}. In \bibinfo{booktitle}{\emph{Proc. of the {ACM} Conference on Computer and Communications Security ({CCS})}}. \bibinfo{pages}{363--376}.
\newblock


\bibitem[Yu et~al\mbox{.}(2020)]%
        {YuRuiQiy20}
\bibfield{author}{\bibinfo{person}{Zeping Yu}, \bibinfo{person}{Rui Cao}, \bibinfo{person}{Qiyi Tang}, \bibinfo{person}{Sen Nie}, \bibinfo{person}{Junzhou Huang}, {and} \bibinfo{person}{Shi Wu}.} \bibinfo{year}{2020}\natexlab{}.
\newblock \showarticletitle{Order matters: Semantic-aware neural networks for binary code similarity detection}. In \bibinfo{booktitle}{\emph{Proc. of the {AAAI} Conference on Artificial Intelligence}}.
\newblock


\bibitem[Zuo et~al\mbox{.}(2019)]%
        {ZuoLiYouLuo+19}
\bibfield{author}{\bibinfo{person}{Fei Zuo}, \bibinfo{person}{Xiaopeng Li}, \bibinfo{person}{Patrick Young}, \bibinfo{person}{Lannan Luo}, \bibinfo{person}{Qiang Zeng}, {and} \bibinfo{person}{Zhexin Zhang}.} \bibinfo{year}{2019}\natexlab{}.
\newblock \showarticletitle{Neural Machine Translation Inspired Binary Code Similarity Comparison beyond Function Pairs.}. In \bibinfo{booktitle}{\emph{Proc. of the Network and Distributed System Security Symposium ({NDSS})}}.
\newblock


\end{thebibliography}

\appendix

\section{Supplementary Results}

\subsection{Baseline Experiment}
\label{sec:appendix:baseline}

\begin{figure}[htbp]
  \centering \begin{tikzpicture}
\matrix (magic) [matrix of nodes, nodes={anchor=east}, draw]
{
\qty{100.00}{\percent} &   \qty{0.00}{\percent} &   \qty{0.00}{\percent} &   \qty{0.00}{\percent} \\
  \qty{0.00}{\percent} & \qty{100.00}{\percent} &  \qty{18.78}{\percent} &  \qty{17.74}{\percent} \\
  \qty{0.00}{\percent} &  \qty{19.23}{\percent} & \qty{100.00}{\percent} &  \qty{69.31}{\percent} \\
  \qty{0.00}{\percent} &  \qty{18.41}{\percent} &  \qty{71.15}{\percent} & \qty{100.00}{\percent} \\
};
\node[font=\ttfamily, xshift=-2mm,left] (magic-1-0) at (magic.west|-magic-1-1) {O0};
\draw[-latex] (magic-1-0) -- (magic-1-1);
\node[font=\ttfamily, xshift=-2mm,left] (magic-2-0) at (magic.west|-magic-2-1) {O1};
\draw[-latex] (magic-2-0) -- (magic-2-1);
\node[font=\ttfamily, xshift=-2mm,left] (magic-3-0) at (magic.west|-magic-3-1) {O2};
\draw[-latex] (magic-3-0) -- (magic-3-1);
\node[font=\ttfamily, xshift=-2mm,left] (magic-4-0) at (magic.west|-magic-4-1) {O3};
\draw[-latex] (magic-4-0) -- (magic-4-1);

\node[font=\ttfamily, yshift=2mm,above] (magic-0-1) at (magic.north-|magic-1-1) {O0};
\draw[latex-] (magic-0-1) -- (magic-1-1);
\node[font=\ttfamily, yshift=2mm,above] (magic-0-2) at (magic.north-|magic-1-2) {O1};
\draw[latex-] (magic-0-2) -- (magic-1-2);
\node[font=\ttfamily, yshift=2mm,above] (magic-0-3) at (magic.north-|magic-1-3) {O2};
\draw[latex-] (magic-0-3) -- (magic-1-3);
\node[font=\ttfamily, yshift=2mm,above] (magic-0-4) at (magic.north-|magic-1-4) {O3};
\draw[latex-] (magic-0-4) -- (magic-1-4);

\end{tikzpicture}
  \caption{Overlap of functions compiled with different optimization
    levels. The arrows indicate the reading direction.}
  \label{fig:option-labeling}
\end{figure}

\paragraph{A closer look at optimization levels}
The results for task~T2 are worse compared to the other
applications. We find that this decrease is not due to a more complex
downstream task, but due to an inherent problem in determining
optimization levels. To understand this problem, we first recall that
the optimization \texttt{O3} activates the same flags as \texttt{O2}
and a few others. This means that the flags enabled by \texttt{O2} are
a subset of the flags enabled by \texttt{O3}. Unfortunately, not every
function provides sufficient complexity to benefit from the unique
flags of \texttt{O3}. As a result, both options may produce the same
assembly instructions for some functions, making it impossible to
recover to the original optimization level. The same is true for the
other optimization level. However, the option \texttt{O2} adds many
more flags in addition to those added by \texttt{O1} making the
differentiation possible in most cases.

\begin{figure}[htbp]
  \centering \begin{tikzpicture}
\matrix (magic) [matrix of nodes, nodes={anchor=east}, draw]
{
\qty{100.00}{\percent} &   \qty{6.76}{\percent} \\
  \qty{7.99}{\percent} & \qty{100.00}{\percent} \\
};
\node[xshift=-2mm,left] (magic-1-0) at (magic.west|-magic-1-1) {\gcc};
\draw[-latex] (magic-1-0) -- (magic-1-1);
\node[xshift=-2mm,left] (magic-2-0) at (magic.west|-magic-2-1) {\clang};
\draw[-latex] (magic-2-0) -- (magic-2-1);

\node[yshift=2mm,above] (magic-0-1) at (magic.north-|magic-1-1) {\gcc};
\draw[latex-] (magic-0-1) -- (magic-1-1);
\node[yshift=2mm,above] (magic-0-2) at (magic.north-|magic-1-2) {\clang};
\draw[latex-] (magic-0-2) -- (magic-1-2);

\end{tikzpicture}
  \caption{Overlap of functions produced with different compilers. The
    arrows indicate the reading direction.}
  \label{fig:compiler-labeling}
\end{figure}
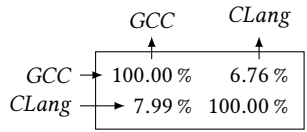

Figure~\ref{fig:option-labeling} shows the overlap between the
different optimization levels in our evaluation corpus. About
\qty{70}{\percent} of the functions compiled with option \texttt{O2}
could be compiled with option \texttt{O3} just as well and vice versa.
Less prevalent are functions compiled with option \texttt{O1} that
could also be compiled with options \texttt{O2} and \texttt{O3}.
Interestingly, the same issue also exists for the compiler-detection
task.  However, in this case it is far less pronounced and the overlap
of the labels is not that common as can be seen in
Figure~\ref{fig:compiler-labeling}.  As a consequence, performance
values as obtained for the other downstream tasks are impossible to
achieve in this setting. Considering this observation, all embedding
types actually produce results close to the optimum for the tasks~T1
and T2.

\begin{figure*}
  \centering
  \input{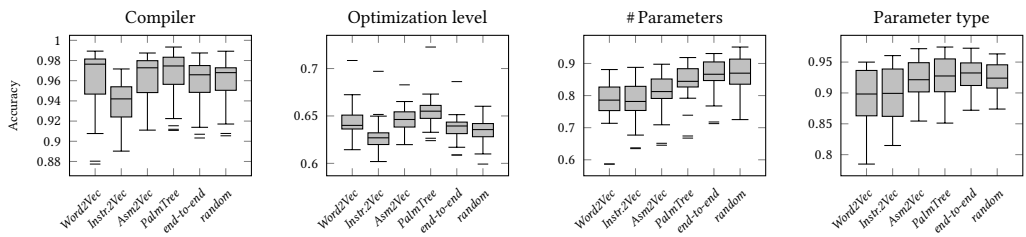}
  \caption{Package accuracy of instances trained on the full dataset
    presented as boxplots for all considered instruction embeddings
    across all binary packages. Outliers are visually highlighted as
    horizontal lines for easy identification.}
  \label{fig:boxplot}
\end{figure*}

\paragraph{Analyzing the package performance}
In addition to the previous observations, we want to check whether
there are significant variations in prediction performance for
different binary packages. For that purpose, we first compute the
package accuracy of each instance.  Second, we compute statistical
characteristics for each downstream task and embedding type.

We depict the results as box-and-whiskers plots in
Figure~\ref{fig:boxplot}. Since we used five different seeds and our
dataset contains \num{30}~different binary packages for testing
(without taking account of different compilation combinations), each
plot is based on \num{150} values. The boxes cover the interquartile
range ($\text{IQR}$) from the first to the third quartile. That means
each box encompasses the mean accuracy values of \qty{50}{\percent} of
the binary packages. The plot shows that the prediction accuracy on
\qty{50}{\percent} of the binary packages differs by less than
\qty{5}{\percent} for tasks T1 and T2 and less than \qty{10}{\percent}
for tasks T3 and T4.
The whiskers extend no more than $1.5 \cdot \text{IQR}$.

Overall, we observe that the majority of the packages fall within a
normal range of packet accuracy. Nevertheless, there are some outliers
that allow for better or worse performance on some of the downstream
tasks. We investigate these outliers by identifying the individual
Debian packages. We find that these packages are outliers for all
considered embedding, indicating again that the choice of an embedding
is less relevant to the underlying learning task when sufficient
labeled data is available.

\subsection{Experiment 1: Size of Labeled Data}
\label{sec:appendix:exp1}

\begin{figure*}
  \centering
  \input{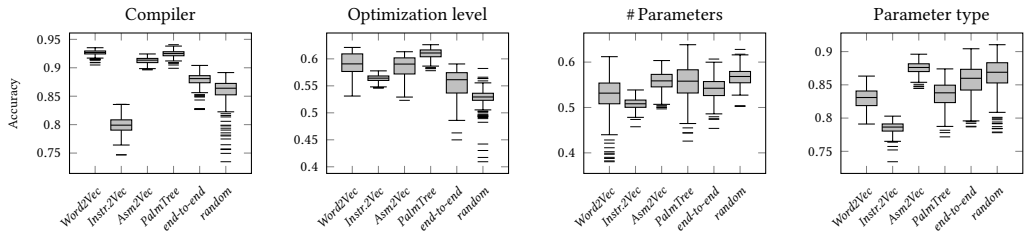}
  \caption{Package accuracy of instances trained with a minimum of
    labeled data presented as boxplots for all considered instruction
    embeddings across all binary packages. Outliers are visually
    highlighted as horizontal lines for easy identification.}
  \label{fig:boxplot_shards256}
\end{figure*}

To check for variations in the general performance of instances
trained on shards, we again use box-and-whiskers plots. Each plot is
based on 256~different instances, each trained on a different data
shard. We show the results of this experiment in
Figure~\ref{fig:boxplot_shards256}. While some variations exist, most
instances produce similar results. We reason that the functions from
some shards are not representative for the test data and no viable
instance can be trained from those functions alone.

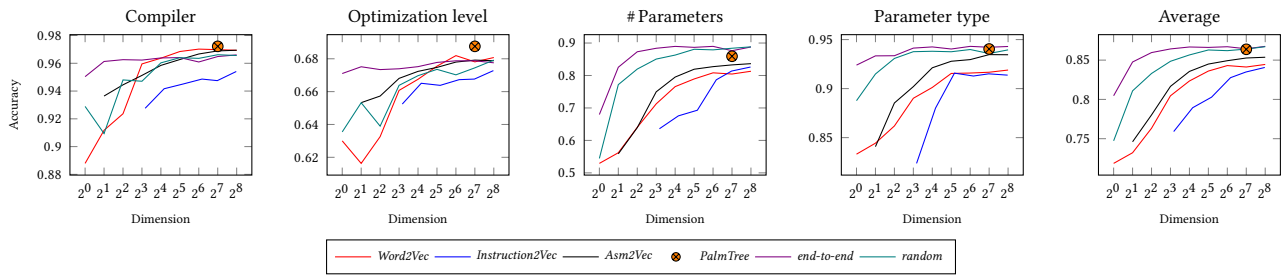
\begin{figure*}
  \centering \begin{tikzpicture}
    \begin{groupplot}[
        tiny,
        xmode=log, log basis x=2,
        group style={group size=5 by 1, xlabels at=edge bottom, ylabels at=edge left},
        xlabel=Dimension,
        ylabel=Accuracy,
        legend columns=-1,
        legend to name=dimension-chart-legend,
    ]

        \nextgroupplot[title=\titlestrut Compiler]

        \addplot+ coordinates {(1, 0.8879113250684756) (2, 0.9117551630209237) (4, 0.9236529363384618) (8, 0.9595424397678999) (16, 0.9636653109885384) (32, 0.9684748243780607) (64, 0.9702102935961185) (128, 0.9
698118902277992) (256, 0.9693576500237684)};
        \addplot+ coordinates {(9, 0.9276022606372945) (18, 0.9416625795108995) (36, 0.9450731538003003) (72, 0.948631620249153) (126, 0.9475118653275887) (252, 0.9540825026975228)};
        \addplot+ coordinates {(2, 0.9364063714356857) (4, 0.9444876215771643) (8, 0.9512695334606011) (16, 0.958489085407722) (32, 0.9626994846410974) (64, 0.9667016275683058) (128, 0.9688943551977303) (256, 0.
9694059413411404)};
        \addplot+[opacity=0.0, mark=otimes*, mark options={draw=black, opacity=1.0}] coordinates {(128, 0.9721872194010368)};
        \addplot+ coordinates {(1, 0.9503776531928861) (2, 0.9613050728519795) (4, 0.9626029020063533) (8, 0.962314663205789) (16, 0.9638448943250156) (32, 0.964108987466894) (64, 0.9609474152826929) (128, 0.964
8605210934965) (256, 0.9660557311984547)};
        \addplot+ coordinates {(1, 0.9289363082796973) (2, 0.9094130341283795) (4, 0.948021942367331) (8, 0.9470606433308936) (16, 0.9603135917421848) (32, 0.9640169321431535) (64, 0.9635430735914403) (128, 0.96
61704230772132) (256, 0.9655969636834202)};

        \nextgroupplot[title=\titlestrut Optimization level]

        \addplot+ coordinates {(1, 0.6300175810577308) (2, 0.6162907740947264) (4, 0.6325785299821172) (8, 0.6607580227723744) (16, 0.6675067343751179) (32, 0.6755019656075274) (64, 0.6819171652996703) (128, 0.6
781142240566216) (256, 0.6807627009937448)};
        \addplot+ coordinates {(9, 0.6524307887330321) (18, 0.6650816047808404) (36, 0.6638139576998242) (72, 0.6672803688249365) (126, 0.667639535497891) (252, 0.6727901063163534)};
        \addplot+ coordinates {(2, 0.6532819232017143) (4, 0.6572991571656015) (8, 0.6680696300432358) (16, 0.6723328479049868) (32, 0.6747459046699213) (64, 0.6779934957631915) (128, 0.679196251386489) (256, 0.
678805393536509)};
        \addplot+[opacity=0.0, mark=otimes*, mark options={draw=black, opacity=1.0}] coordinates {(128, 0.687564231224864)};
        \addplot+ coordinates {(1, 0.6710199277139343) (2, 0.6751352534162335) (4, 0.6734163843385221) (8, 0.6739128794452535) (16, 0.6752076903922914) (32, 0.6777761848350172) (64, 0.6787631386338084) (128, 0.6
787601204264727) (256, 0.6775407646628285)};
        \addplot+ coordinates {(1, 0.6354578997804254) (2, 0.6532275954696708) (4, 0.6389861841559206) (8, 0.6638260305291672) (16, 0.6697085166265496) (32, 0.673552203668631) (64, 0.6702201027699598) (128, 0.67
44833206317108) (256, 0.6791871967644817)};

        \nextgroupplot[title=\titlestrut \#\,Parameters]

        \addplot+ coordinates {(1, 0.529210838951821) (2, 0.5619758915802139) (4, 0.6402760436766475) (8, 0.7126073566245186) (16, 0.7656927954719152) (32, 0.7892639591988074) (64, 0.8082071538292737) (128, 0.80
44902499375723) (256, 0.8130515387467556)};
        \addplot+ coordinates {(9, 0.6359507540502296) (18, 0.6756274923762609) (36, 0.6929710260077334) (72, 0.7866866435116872) (126, 0.8136357101238716) (252, 0.8260894569173609)};
        \addplot+ coordinates {(2, 0.5586191762578224) (4, 0.6394376215447247) (8, 0.7497143462501797) (16, 0.7951601552745681) (32, 0.8192564678819247) (64, 0.8272487192874926) (128, 0.8326409540456895) (256, 0
.8364607689571936)};
        \addplot+[opacity=0.0, mark=otimes*, mark options={draw=black, opacity=1.0}] coordinates {(128, 0.8592313454859141)};
        \addplot+ coordinates {(1, 0.679743933168373) (2, 0.8254235620833428) (4, 0.8730622838679409) (8, 0.883852806973735) (16, 0.8896445786323428) (32, 0.8866828600183121) (64, 0.8897368958706954) (128, 0.876
3841910512815) (256, 0.8878148812361429)};
        \addplot+ coordinates {(1, 0.544511286160738) (2, 0.7718311351236824) (4, 0.8197906971464893) (8, 0.8505868198224785) (16, 0.8629073876491642) (32, 0.8807518558035005) (64, 0.8789524263543015) (128, 0.88
3963284980288) (256, 0.8882446860835547)};

        \nextgroupplot[title=\titlestrut Parameter type]

        \addplot+ coordinates {(1, 0.8331205401427154) (2, 0.8443377811910019) (4, 0.8618844196985697) (8, 0.8901491152283127) (16, 0.9011775623180058) (32, 0.9154683050142396) (64, 0.915873092061054) (128, 0.91
66986656426994) (256, 0.9189337941185882)};
        \addplot+ coordinates {(9, 0.8237160410866853) (18, 0.880096636907619) (36, 0.9157658954913442) (72, 0.9128347892867428) (126, 0.9150667178650284) (252, 0.9137339605132636)};
        \addplot+ coordinates {(2, 0.8408002943905795) (4, 0.8853396691305878) (8, 0.9020239352340725) (16, 0.9212409202905507) (32, 0.927888707561358) (64, 0.9295846532910946) (128, 0.9346628907874948) (256, 0.
9345348948833637)};
        \addplot+[opacity=0.0, mark=otimes*, mark options={draw=black, opacity=1.0}] coordinates {(128, 0.9403859076509552)};
        \addplot+ coordinates {(1, 0.9239544334581293) (2, 0.9333893315413907) (4, 0.9335381267799431) (8, 0.941315477904707) (16, 0.9425266391475473) (32, 0.9403859076509552) (64, 0.9430882211769224) (128, 0.94
21810502063934) (256, 0.9429602252727912)};
        \addplot+ coordinates {(1, 0.887683594124988) (2, 0.9148891235480464) (4, 0.9307478160698858) (8, 0.9376531950977569) (16, 0.9380867812230008) (32, 0.9375955969408979) (64, 0.9398627243928195) (128, 0.93
58052542318646) (256, 0.939505935810054)};

        \nextgroupplot[title=\titlestrut Average]

        \addplot+ coordinates {(1, 0.7185733845255307) (2, 0.7321176143925394) (4, 0.7632856199075093) (8, 0.8046153404249208) (16, 0.8234481966239958) (32, 0.8360836054115642) (64, 0.8430428080388346) (128, 0.8
412335915283935) (256, 0.8444920195704835)};
        \addplot+ coordinates {(9, 0.7590950087607594) (18, 0.7894104918472277) (36, 0.8028815717664181) (72, 0.8276786381088249) (126, 0.8348396733146309) (252, 0.8406468600097661)};
        \addplot+ coordinates {(2, 0.7460651264325996) (4, 0.7802491311050046) (8, 0.816604272186743) (16, 0.8356190457953142) (32, 0.8449891328284327) (64, 0.8492574897312409) (128, 0.8526993671189069) (256, 0.
8536660187853662)};
        \addplot+[opacity=0.0, mark=otimes*, mark options={draw=black, opacity=1.0}] coordinates {(128, 0.8637576477121492)};
        \addplot+ coordinates {(1, 0.8046693411717396) (2, 0.8476114244133164) (4, 0.8595957603668987) (8, 0.8642409735453788) (16, 0.8667135183784457) (32, 0.8661704470380973) (64, 0.8670382888273986) (128, 0.8
644343804754746) (256, 0.8675074442518886)};
        \addplot+ coordinates {(1, 0.7472983349769731) (2, 0.8108916825445458) (4, 0.8330086267150503) (8, 0.8485148837164769) (16, 0.8565927826661432) (32, 0.8629062742357075) (64, 0.8620277088938464) (128, 0.8
640732265446224) (256, 0.8670907576381376)};

        \legend{\wtov, \itov, \atov, \palmtree, \etoe, \rand};

    \end{groupplot}
    \node[below] at (current bounding box.south) {\pgfplotslegendfromname{dimension-chart-legend}};
\end{tikzpicture}
  \caption{ General accuracy of the considered embeddings in relation
    to the embedding dimension.}
  \label{fig:dimension-chart}
\end{figure*}

\subsection{Experiment 4: Sequence length}
\label{sec:appendix:exp4}

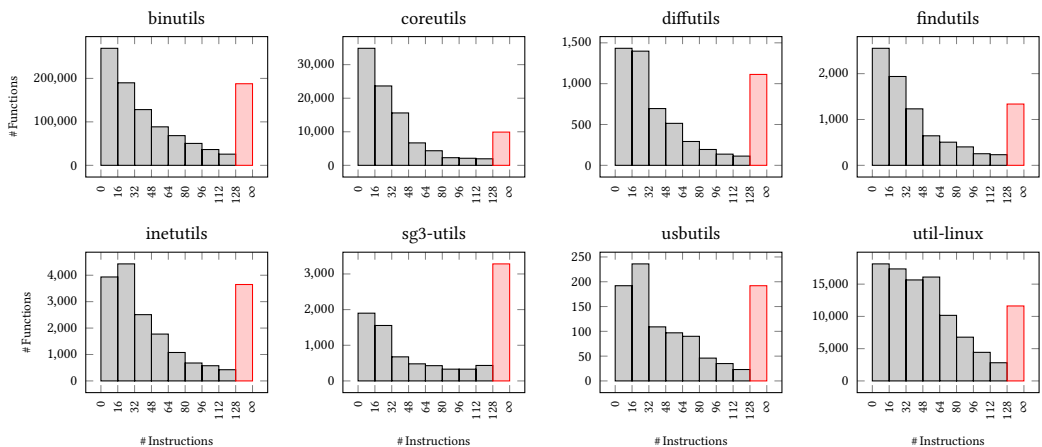
\begin{figure*}
  \centering \begin{tikzpicture}
    \begin{groupplot}[
        tiny,
        scaled ticks=false,
        ymin=0.0,
        enlargelimits=0.1,
        xlabel=\#\,Instructions,
        ylabel=\#\,Functions,
        xtick={0, 16, 32, 48, 64, 80, 96, 112, 128, 144},
        xticklabels={0, 16, 32, 48, 64, 80, 96, 112, 128, $\infty$},
        xticklabel style={rotate=90},
        yticklabel style={/pgf/number format/fixed},
        group style={group size=4 by 2, xlabels at=edge bottom, ylabels at=edge left},
    ]
    \nextgroupplot[title=\mystrut binutils];

    \addplot+[black, ybar interval, fill opacity=0.2, fill] coordinates { (0.0, 269217) (16.0, 189830) (32.0, 128203) (48.0, 88539) (64.0, 68320) (80.0, 50407) (96.0, 36349) (112.0, 25714) (128.0, 187710) };
    \addplot+[red, ybar interval, fill opacity=0.2, fill=red] coordinates { (128.0, 187710) (144.0, 0) };

    \nextgroupplot[title=\mystrut coreutils];

    \addplot+[black, ybar interval, fill opacity=0.2, fill] coordinates { (0.0, 34873) (16.0, 23652) (32.0, 15610) (48.0, 6685) (64.0, 4348) (80.0, 2275) (96.0, 2107) (112.0, 1948) (128.0, 9904) };
    \addplot+[red, ybar interval, fill opacity=0.2, fill] coordinates { (128.0, 9904) (144.0, 0) };

    \nextgroupplot[title=\mystrut diffutils];

    \addplot+[black, ybar interval, fill opacity=0.2, fill] coordinates { (0.0, 1433) (16.0, 1398) (32.0, 695) (48.0, 514) (64.0, 292) (80.0, 194) (96.0, 137) (112.0, 113) (128.0, 1113) };
    \addplot+[red, ybar interval, fill opacity=0.2, fill] coordinates { (128.0, 1113) (144.0, 0) };

    \nextgroupplot[title=\mystrut findutils];

    \addplot+[black, ybar interval, fill opacity=0.2, fill] coordinates { (0.0, 2553) (16.0, 1938) (32.0, 1232) (48.0, 644) (64.0, 505) (80.0, 402) (96.0, 253) (112.0, 232) (128.0, 1337) };
    \addplot+[red, ybar interval, fill opacity=0.2, fill] coordinates { (128.0, 1337) (144.0, 0) };

    \nextgroupplot[title=\mystrut inetutils];

    \addplot+[black, ybar interval, fill opacity=0.2, fill] coordinates { (0.0, 3934) (16.0, 4429) (32.0, 2508) (48.0, 1773) (64.0, 1076) (80.0, 678) (96.0, 575) (112.0, 424) (128.0, 3649) };
    \addplot+[red, ybar interval, fill opacity=0.2, fill] coordinates { (128.0, 3649) (144.0, 0) };

    \nextgroupplot[title=\mystrut sg3-utils];

    \addplot+[black, ybar interval, fill opacity=0.2, fill] coordinates { (0.0, 1900) (16.0, 1555) (32.0, 675) (48.0, 479) (64.0, 428) (80.0, 330) (96.0, 329) (112.0, 434) (128.0, 3282) };
    \addplot+[red, ybar interval, fill opacity=0.2, fill] coordinates { (128.0, 3282) (144.0, 0) };

    \nextgroupplot[title=\mystrut usbutils];

    \addplot+[black, ybar interval, fill opacity=0.2, fill] coordinates { (0.0, 192) (16.0, 236) (32.0, 109) (48.0, 97) (64.0, 90) (80.0, 46) (96.0, 35) (112.0, 23) (128.0, 192) };
    \addplot+[red, ybar interval, fill opacity=0.2, fill] coordinates { (128.0, 192) (144.0, 0) };

    \nextgroupplot[title=\mystrut util-linux];

    \addplot+[black, ybar interval, fill opacity=0.2, fill] coordinates { (0.0, 18137) (16.0, 17358) (32.0, 15651) (48.0, 16097) (64.0, 10162) (80.0, 6781) (96.0, 4429) (112.0, 2818) (128.0, 11618) };
    \addplot+[red, ybar interval, fill opacity=0.2, fill] coordinates { (128.0, 11618) (144.0, 0) };

    \end{groupplot}
\end{tikzpicture}
  \caption{Distribution of function lengths. The red bar accumulates
    all functions with more than 128 instructions.}
  \label{fig:function-length}
\end{figure*}

\begin{figure}
    \centering
    \begin{tikzpicture}

   \begin{groupplot}[
        tiny,
        xmode=log, log basis x=2,
        group style={group size=4 by 1, xlabels at=edge bottom, ylabels at=edge left},
        xlabel=Sequence~length,
        ylabel=Performance,
        ymin=0.0,
        legend columns=-1,
        legend to name=sequence-length-chart-legend
    ]

    \nextgroupplot[title=\titlestrut Accuracy]

    \addplot+ coordinates {(1, 0.6021449393466084) (2, 0.7108758586171077) (4, 0.8889651824632093) (8, 0.9525235986086065) (16, 0.968007002241019) (32, 0.9723456752861638) (64, 0.9738396879173615) (128, 0.9732813195602472) (256, 0.9731756823034958)};
    \addplot+ coordinates {(1, 0.519252390042934) (2, 0.6097659128693845) (4, 0.6371184168496455) (8, 0.6510725954319432) (16, 0.665074562297057) (32, 0.6804774804005161) (64, 0.6904451101268402) (128, 0.6941801417048344) (256, 0.6972914104334397)};
    \addplot+ coordinates {(1, 0.3314996002108667) (2, 0.39234574066927475) (4, 0.5087436531898633) (8, 0.7553517513790833) (16, 0.8440519700650004) (32, 0.8567846864366807) (64, 0.8600914596465208) (128, 0.8592439571304977) (256, 0.8563861584678365)};
    \addplot+ coordinates {(1, 0.8420210553262295) (2, 0.8510047678474288) (4, 0.8820277751111965) (8, 0.9181332864015018) (16, 0.9247544078589485) (32, 0.9374180026239161) (64, 0.9411592162384137) (128, 0.9429378259895683) (256, 0.9309382099772807)};
    
    \draw[help lines] (128,0) -- (128,1);
    
    \nextgroupplot[title=\titlestrut Balanced Accuracy]
    
    \addplot+ coordinates {(1, 0.6039041721825755) (2, 0.71198185001356) (4, 0.8897102539088961) (8, 0.9528960598001974) (16, 0.9682380045419942) (32, 0.9724913517437928) (64, 0.9739371070919863) (128, 0.9732453415191193) (256, 0.9732794469547614)};
    \addplot+ coordinates {(1, 0.4614706495491769) (2, 0.5526449432457295) (4, 0.5856177427450796) (8, 0.6008680622294706) (16, 0.616511473769444) (32, 0.6355689780368663) (64, 0.6466009169756476) (128, 0.652467662110702) (256, 0.6539117687759541)};
    \addplot+ coordinates {(1, 0.1324829407927819) (2, 0.1870429309460706) (4, 0.29970660429430557) (8, 0.5077153666679151) (16, 0.640111593327516) (32, 0.6578900492434527) (64, 0.6599434484140828) (128, 0.6519533610883822) (256, 0.6445748635426775)};
    \addplot+ coordinates {(1, 0.26045792592339556) (2, 0.2613042959632341) (4, 0.4555084188221402) (8, 0.5123917598140703) (16, 0.6034043884501029) (32, 0.5724602411572419) (64, 0.599952755237268) (128, 0.6276402061383073) (256, 0.5521734948410301)};
    
    \draw[help lines] (128,0) -- (128,1);

    \legend{Compiler, Optimization level, \#\,Parameters, Parameter type};
    \end{groupplot}
\end{tikzpicture}
    \caption{General accuracy for the \palmtree embedding in relation
      to the sequence length.}
    \label{fig:seq-len}
\end{figure}
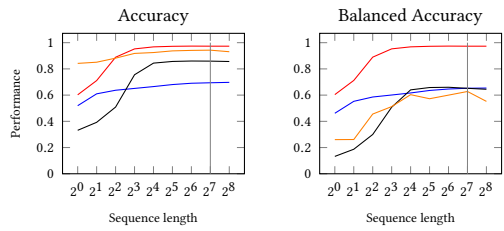

Figure~\ref{fig:seq-len} shows the results of this experiment. It
shows the general accuracy of instances trained with the \palmtree
instruction embedding model in relation to the sequence length. That
is, the maximum number of instructions of each function used to train
the instances. The plot clearly shows that a certain number of
instructions is needed. This comes at no surprise, however, the plot
also shows that after processing about \num{16} instructions the
performance starts to plateaus and only little gain can be
observed. We observe no further improvement after \num{128}
instructions. Further, Figure~\ref{fig:function-length} shows that
only few functions are longer than \num{256} instructions. Because of
that, we limit the instruction sequences in the other experiments to
this threshold.

However, that such short sequences are sufficient raises questions, at
least for tasks~T3 and T4, since we expect that necessary information
is missing in such short sequences. For that reason we measure the
general accuracy with a different metric, the balanced accuracy, which
takes the label imbalance into account. The results are also shown in
Figure~\ref{fig:seq-len} and gives a different picture. The reason is,
that now the underrepresented type categories have more weight. The
classes of tasks~T1 and T2 are more balanced and, hence, the
performance is not affected.

\begin{table}
  \caption{General balanced accuracy of the considered embeddings
    witha minimum of labeled data. The second value (\textdownarrow)
    displayed represents the margin to the balanced accuracy achieved
    on the full dataset.}
  \label{tab:bal_accuracy_drop}
  \centering 
%

\setlength{\tabcolsep}{4pt}
\begin{tabular}{lrrrrr}
\toprule
 & Task T1 & Task T2 & Task T3 & Task T4 & Average \\
  \addlinespace
  \multicolumn{6}{l}{Instruction embeddings:} \\
  \midrule
\wtov     & $0.94_{\downarrow .02}$ & $0.59_{\downarrow .04}$ & $0.28_{\downarrow .32}$ & $0.26_{\downarrow .28}$ & $0.52_{\downarrow .17}$ \\
\itovx    & $0.83_{\downarrow .12}$ & $0.57_{\downarrow .06}$ & $0.30_{\downarrow .33}$ & $0.23_{\downarrow .25}$ & $0.48_{\downarrow .19}$ \\
\atov     & $0.93_{\downarrow .04}$ & $0.59_{\downarrow .05}$ & $0.31_{\downarrow .33}$ & $0.31_{\downarrow .32}$ & $0.53_{\downarrow .19}$ \\
\palmtree & $0.94_{\downarrow .03}$ & $0.61_{\downarrow .04}$ & $0.31_{\downarrow .32}$ & $0.29_{\downarrow .31}$ & $0.54_{\downarrow .18}$ \\
\etoe     & $0.89_{\downarrow .07}$ & $0.56_{\downarrow .09}$ & $0.27_{\downarrow .39}$ & $0.29_{\downarrow .30}$ & $0.50_{\downarrow .21}$ \\
\rand     & $0.87_{\downarrow .09}$ & $0.53_{\downarrow .10}$ & $0.30_{\downarrow .41}$ & $0.30_{\downarrow .19}$ & $0.50_{\downarrow .20}$ \\
  \addlinespace
  \multicolumn{6}{l}{Function embeddings:} \\
  \midrule
\atov     & $0.86_{\downarrow .07}$ & $0.57_{\downarrow .04}$ & $0.19_{\downarrow .06}$ & $0.27_{\downarrow .20}$ & $0.47_{\downarrow .09}$ \\
\ada      & $0.70_{\downarrow .17}$ & $0.44_{\downarrow .18}$ & $0.11_{\downarrow .27}$ & $0.20_{\downarrow .42}$ & $0.36_{\downarrow .26}$ \\
\bottomrule
\end{tabular}
\end{table}

Table~\ref{tab:bal_accuracy_drop} serves the same purpose as
Table~\ref{tab:accuracy_drop}, but shows the performance based on the
balanced accuracy. The average performance is noticeable
lower. Moreover, the performance drop, that is, the margin to the
performance obtained on the full dataset, is significant.

\end{document}